\def\year{2021}\relax
\documentclass[letterpaper]{article} % DO NOT CHANGE THIS
\usepackage{aaai21}  % DO NOT CHANGE THIS
\usepackage{times}  % DO NOT CHANGE THIS
\usepackage{helvet} % DO NOT CHANGE THIS
\usepackage{courier}  % DO NOT CHANGE THIS
\usepackage[hyphens]{url}  % DO NOT CHANGE THIS
\usepackage{graphicx} % DO NOT CHANGE THIS
\usepackage{natbib}  % DO NOT CHANGE THIS AND DO NOT ADD ANY OPTIONS TO IT
\usepackage{caption} % DO NOT CHANGE THIS AND DO NOT ADD ANY OPTIONS TO IT
\usepackage{hyperref}
\usepackage{fullpage}
\usepackage{subfig}
\usepackage{amsmath}
\usepackage{amssymb}
\usepackage{amsthm}
\usepackage{algorithm}
\usepackage{algpseudocode}
\usepackage{amsfonts}
\usepackage{tikz}
\usepackage{enumitem}
\usepackage{tabularx}
\usepackage{booktabs}

\usetikzlibrary{matrix}
\usetikzlibrary{shapes,arrows}
\usetikzlibrary{positioning}

\usepackage{pgfplots}
\pgfplotsset{width=5cm,compat=1.12}
\usepgfplotslibrary{fillbetween}

\makeatletter
\newtheorem*{rep@theorem}{\rep@title}
\newcommand{\newreptheorem}[2]{%
\newenvironment{rep#1}[1]{%
 \def\rep@title{#2 \ref{##1}}%
 \begin{rep@theorem}}%
 {\end{rep@theorem}}}
\makeatother

\newtheorem{definition}{Definition}
\newtheorem{theorem}{Theorem}
\newtheorem{proposition}{Proposition}
\newtheorem{corollary}{Corollary}
\newtheorem{lemma}{Lemma}
\newtheorem{assumption}{Assumption}
\newtheorem{remark}{Remark}

\newreptheorem{proposition}{Proposition}
\newreptheorem{corollary}{Corollary}
\newreptheorem{theorem}{Theorem}
\newreptheorem{lemma}{Lemma}  
\newreptheorem{definition}{Definition}

\title{A Theoretical Analysis of Provable Compositional Generalization \\ in Neural Networks: A Necessary and Sufficient Condition}
\author{
    Yuanpeng Li\thanks{\texttt{yuanpeng16@foxmail.com}}
}
\begin{document}

\maketitle

\begin{abstract}
Compositional generalization---the ability to systematically process novel combinations of known components---is a hallmark of human intelligence; however, its theoretical foundation in neural networks is not yet well understood. This paper establishes a necessary and sufficient condition for provable compositional generalization, precisely characterizing its boundary. Conceptually, the condition consists of two principles: (i)~structural alignment, where a model's computational graph aligns with a task's true compositional hierarchy, and (ii)~unambiguous minimized representations, where each component encodes adequate but not redundant information on the training data. The result is fully proved and machine-verified in Lean~4 and holds even in few-shot and one-shot regimes. The necessity direction establishes that provable compositional generalization cannot circumvent these requirements, while the sufficiency direction yields a unified inductive bias that jointly governs architectural design, training data properties, and regularization strategies. Building on this condition, we develop an example algorithmic approach, illustrate it through a controlled minimal example, and further demonstrate the condition on the SCAN jump task. All conclusions are derived mathematically without reliance on empirical validation. Our work provides a theoretical characterization of provable compositional generalization.
Resources are available at \url{https://github.com/yuanpeng16/tacg}.
\end{abstract}

\section{Introduction}
Compositional generalization~\cite{fodor1988connectionism,lake2018generalization} is a cornerstone of human intelligence, endowing cognitive systems with the algebraic capacity to understand and produce a potentially infinite number of novel combinations from a finite set of known components~\cite{chomsky1957syntactic,montague1970universal}. In natural language processing, this capability enables humans to interpret sentences they have never encountered before, so long as the words and grammatical rules are familiar~\cite{kim-linzen-2020-cogs,keysers2020measuring}. In computer vision, it supports recognition of previously unseen object arrangements from knowledge of individual features~\cite{andreas2016neural,higgins2017beta}.

Despite its importance, the theoretical foundation of compositional generalization 
in neural networks is not yet well understood. In this work, we derive a necessary and sufficient condition for \textit{provable} compositional generalization, thereby precisely characterizing its theoretical boundary.

Our analysis is motivated by the findings of \citet{li2019compositional}, which showed that structural alignment and representation compression together enable compositional generalization for primitive substitution tasks. We first establish necessity under a few reasonable assumptions, and then prove sufficiency without them. Through precise formal definitions and rigorous mathematical analysis, we obtain our main result:

\begin{theorem}[Necessary and Sufficient Condition]
\label{theorem:necessary_and_sufficient_condition}
A model provably enables compositional generalization (Definition~\ref{definition:compositional_generalization}) if and only if it has Aligned Structure-Unambiguous Minimized Representation (AS-UMR, Definition~\ref{definition:aligned_structure-unambiguous_minimized_representation}).

\end{theorem}

Conceptually, it says that a model provably generalizes compositionally if and only if it satisfies two principles:
\begin{enumerate}[label=(\roman*),leftmargin=0.65cm]
    \item \textbf{Structural Alignment}: A model's computational graph aligns with a task's true compositional hierarchy.
    \item \textbf{Unambiguous Minimized Representations}: Each component encodes adequate but not excessive information on the training data, i.e., no ambiguities and no redundancies.
\end{enumerate}

Our main contributions are as follows:
\begin{enumerate}[leftmargin=0.65cm]
    \item A \textit{necessary and sufficient condition} (Theorem~\ref{theorem:necessary_and_sufficient_condition}) for provable compositional generalization, fully proved mathematically and formally verified in the Lean~4 theorem prover~\cite{10.1007/978-3-030-79876-5_37}. It holds even in challenging few-shot and one-shot learning regimes.
    \item A unified \textit{inductive bias}~\cite{goyal2022inductive}, which jointly governs architectural design (structural alignment), training data properties (unambiguous representation), and regularization strategies (minimized representation).
    \item An \textit{example approach} based on the condition, illustrated through a carefully designed minimal example. We further demonstrate the condition on the SCAN jump task (Appendix~\ref{sec:scan_jump_task}) and discuss it in the context of the COGS task (Appendix~\ref{section:cogs_task}).
\end{enumerate}

This work is purely theoretical: all conclusions are established by mathematical proofs and do not depend on empirical validation.

Throughout this paper, we assume that standard in-domain generalization (i.e., all test inputs lie within the support of the training distribution) is already addressed, and we focus on the core problem of compositional generalization, where test samples are novel combinations of known components.

Before formalizing the framework, we preview a concrete example. Consider a three-input XOR task defined by the function \(y=(x_1\oplus x_2)\oplus x_3\), with each \(x_i \in \{0, 1\}\). Training covers six of the eight input combinations; the unseen ones, \(100\) and \(110\), require compositional generalization. The following conditions enable it:
\begin{enumerate}
    \item \textbf{Structural alignment}: the network mirrors the true two-step hierarchy, i.e., first compute \(z=x_1\oplus x_2\), then \(y=z\oplus x_3\).
    \item For training data, the node \(h\) in the model (corresponding to \(z\)) has:
    \begin{enumerate}
        \item \textbf{Unambiguous representation}: the value of \(h\) uniquely determines whether the true intermediate value \(z\) is \(0\) or \(1\).
        \item \textbf{Minimized representation}: the number of distinct values of \(h\) is exactly two, matching the number of possible values of \(z\), i.e., \(|\{0,1\}|=2\).
    \end{enumerate}
\end{enumerate}
Further details are given in Section~\ref{sec:minimal_example}.

\section{Definitions and Assumptions}
\label{section:definitions_and_assumptions}
We establish the formal definitions and assumptions required for our theoretical analysis. All sets considered in this paper are finite. A complete list of symbols appears in Appendix~\ref{section:list_of_symbols}.

\subsection{Setting Definitions}
Let $\mathcal{D}_{\mathrm{train}}$ denote the training dataset and $\mathcal{D}_{\mathrm{test}}$ the test dataset. The full dataset is $\mathcal{D} = \mathcal{D}_{\mathrm{train}} \cup \mathcal{D}_{\mathrm{test}}$. Each sample is an input-output pair $(X, Y)$.

\begin{definition}[Component]
\label{definition:component}
A \textit{component} is a deterministic function that maps a tuple of input nodes to a single output node. A component is either an \textit{output component} or a \textit{non-output component}.
\end{definition}

The definition of a component is intentionally broad: it only specifies the input-output signature and determinism; any additional information is regarded as inductive bias.
A node may be multi-dimensional. Neural network modules qualify as components because they behave deterministically at inference time. Note that stochastic training mechanisms such as dropout or noise injection do not violate inference-time determinism.

\begin{definition}[Computational Graph]
\label{definition:computational_graph}
A \textit{computational graph} is a directed acyclic graph that maps a sample's input to its output by composing components, satisfying:
\begin{itemize}
    \item For each sample $(X, Y)$, the graph input is $X$, and the graph output $\hat{Y}$ has the same number of nodes as $Y$.
    \item Graph input nodes and graph output nodes are disjoint.
    \item Graph output nodes have no outgoing edges, and graph input nodes have no incoming edges.
    \item Every non-output node is an input to at least one non-input node (no dangling nodes).
    \item The graph has a fixed topological order (computational order). If multiple topological orders exist, any one suffices.
    \item In the topological order, each step applies a component to a tuple of previous node values (as specified by the node's input indices) to produce a new node value.
    \item A node is a graph output node if and only if it is the output of an output component.
\end{itemize}
\end{definition}

We refer to computational graphs simply as \textit{graphs}. Graphs are not provided in the data. Components may be reused multiple times within a graph: for instance, a convolution kernel applied at different receptive fields corresponds to the same component instantiated at different graph positions, and a recurrent network reuses one component across timesteps.

A model produces a \textit{hypothesis graph} $H$ for each sample. By a slight abuse of notation, $H$ also denotes its node set. Then, $H \setminus X$ denotes the non-input nodes. For a node $h \in H$, we write $\mathbf{h}$ for its tuple of input nodes, so that $(\mathbf{h}, h)$ denotes one specific invocation of a component. The arity $n = |\mathbf{h}|$ may vary across components. The $i$-th input node is $h_i \in \mathbf{h}$, and $(h_i, h)$ is the corresponding directed edge.

Let $c(h)$ be the component computing node $h$ (with $c(h) = \epsilon$ for input nodes, where $\epsilon$ is a null component identifier), and let $v(h)$ be its computed value. Note that comparisons are meaningful only within the same component; hence, when the context is clear, we write $h = h'$ to mean $c(h) = c(h')$ and $v(h) = v(h')$. Similarly, $\mathbf{h} = \mathbf{h}'$ means both tuples feed the same component and $h_i = h'_i$ for all $i$.

Throughout, when we say two representations are equal, e.g., $v(h) = v(h')$, we mean that they are close enough to be indistinguishable under in-domain generalization, not necessarily numerically identical.

\begin{definition}[Graph Set]
A \textit{graph set} is a collection of graphs, one per sample in the full dataset.
\end{definition}

The \textit{non-output component set} of a graph set is the collection of all non-output components used across its graphs. A \textit{hypothesis graph set} $\mathcal{H}$ is the graph set produced by a model. We index samples by $A, B, C, \dots$, with $A, C, \dots$ denoting training samples and $B, D, \dots$ denoting test samples by default.

\begin{definition}[Reference Graph Set]
\label{definition:reference_graph_set}
A \textit{reference graph set} $\mathcal{Z}$ is a graph set satisfying:
\begin{enumerate}
    \item \textit{Correct predictions on all samples}:
    \[
    \forall A \in \mathcal{D}: \hat{Y}^A = Y^A.
    \]
    \item \textit{All test component inputs appear in training}:
    \begin{align*}
    & \forall B \in \mathcal{D}_{\mathrm{test}}, \forall z^B \in Z^B \setminus X^B, \; \\
    & \exists A \in \mathcal{D}_{\mathrm{train}}, \exists z^A \in Z^A \setminus X^A: \mathbf{z}^A = \mathbf{z}^B,
    \end{align*}
    where $\mathbf{z}^A$ and $\mathbf{z}^B$ are input tuples for the same component, with $z^A_i = z^B_i$ component-wise.
\end{enumerate}
\end{definition}

The second condition captures the essence of compositional generalization: every component-level input encountered at test time has been observed during training, even though the model-level input is novel.

Let $\mathbb{Z}$ be the set of all reference graph sets compatible with a given train/test split. Reference graph structures may differ across samples. Figure~\ref{fig:an_example_of_notation} illustrates our notation.

\begin{figure}[!ht]
    \centering
    \begin{tikzpicture}[xscale=1.2,yscale=0.85]
    \path [every node]
      node (layer1) at (0,0) {$h$}
      node (layer6) at (0,-0.5) {$\uparrow$}
      node (layer2) at (0,-1) {$h_1, \dots, h_n$}
      node (layer4) at (-1,-1) {$\mathbf{h} =$}
      node (layer3) at (0,-2) {$\mathbf{h}_1, \dots, \mathbf{h}_n$}
      node (layer5) at (0,-1.5) {$\uparrow \quad\quad\quad \uparrow$}
      node (hlayer1) at (3,0) {$z$}
      node (hlayer6) at (3,-0.5) {$\uparrow$}
      node (hlayer2) at (3,-1) {$z_1, \dots, z_n$}
      node (hlayer4) at (2,-1) {$\mathbf{z} =$}
      node (hlayer3) at (3,-2) {$\mathbf{z}_1, \dots, \mathbf{z}_n$}
      node (hlayer5) at (3,-1.5) {$\uparrow \quad\quad\quad \uparrow$}
      ;
    \end{tikzpicture}
    \caption{
        An example of notation.
        Normal fonts (e.g., $h, h_1$)  denote nodes, and bold fonts (e.g., $\mathbf{h}, \mathbf{h}_1$) denote the corresponding input nodes.
        An arrow points from the input nodes to the output node of a component.
        On the left, $(\mathbf{h}, h)$ and $\forall i \in \{1, \dots, n\}: (\mathbf{h}_i, h_i)$ represent specific uses of components. Similarly on the right.
    }
    \label{fig:an_example_of_notation}
\end{figure}

We state all mappings from hypothesis to reference by default, as predictions are naturally compared against their ground-truth. The choice of direction is purely notational and does not affect the results.

\subsection{Condition Definitions}
\begin{definition}[Compositional Generalization]
\label{definition:compositional_generalization}
A model \textit{enables compositional generalization} if correct predictions on all training samples imply correct predictions on all test samples:
\begin{align*}
    \text{If} \quad & \forall A \in \mathcal{D}_{\mathrm{train}}: \hat{Y}^A = Y^A, \\
    \text{then} \quad & \forall B \in \mathcal{D}_{\mathrm{test}}: \hat{Y}^B = Y^B.
\end{align*}
\end{definition}

This definition does not require the training support to cover the test one. The compositional aspect---recombining seen components into unseen wholes---is encoded in the reference graph set requirement (Definition~\ref{definition:reference_graph_set}), where all test component inputs have been observed in training. Section~\ref{sec:alternative_definition} presents an alternative definition that bakes this requirement in directly.

\begin{definition}[Structural Alignment]
\label{definition:structure_alignment}
$\mathcal{H}$ and $\mathcal{Z}$ have \textit{structural alignment}, written $\mathcal{H} \cong_S \mathcal{Z}$, if there exists a bijection $\varphi$ between the components of $\mathcal{H}$ and those of $\mathcal{Z}$ (including output components on both sides) such that for every sample $A \in \mathcal{D}$:
\begin{enumerate}
    \item $H^A$ and $Z^A$ contain the same number of nodes; and
    \item for each pair of nodes $h^A \in H^A$ and $z^A \in Z^A$ that occupy the same position in their respective topological orders:
    \begin{enumerate}
        \item the corresponding components are aligned, i.e., $\varphi: c(h^A) \mapsto c(z^A)$, and
        \item their input index tuples are identical.
    \end{enumerate}
\end{enumerate}
\end{definition}

\begin{remark}[Structure as a Program]
\label{remark:structure_as_a_program}
The required structure need not be sample-specific (see Appendix~\ref{sec:scan_jump_task} for an example). It can correspond to a single uniform program that handles the entire task---for instance, a chart parser applied to every input, rather than a per-sample parse tree.
\end{remark}

Structural alignment requires only equal graph topology. Given the alignment, there is a canonical bijection between nodes of $H^A$ and $Z^A$ (same topological index), so we may treat the corresponding nodes $h^A$ and $z^A$ interchangeably. Because $\varphi$ is a bijection, we identify the \textbf{non-output} component sets of $\mathcal{H}$ and $\mathcal{Z}$ and denote them by $\mathcal{C}$.

\begin{definition}[Component-wise Value Mapping]
\label{definition:component-wise_value_mapping}
Given $\mathcal{H} \cong_S \mathcal{Z}$ (Definition~\ref{definition:structure_alignment}), for each $c \in \mathcal{C}$, define the mapping $\xi_c$ as the set of pairs $(v(h), v(z))$ for which the nodes $h$ and $z$ have component $c$ and appear at the same topological index in some training sample.
\end{definition}

We use ``mapping'' in the set-theoretic sense: a mapping need not be single-valued. A \textit{function} is a single-valued mapping. In general, $\xi_c$ may not be well-defined (i.e., may not be a function). By construction, every reference value appears in at least one training sample, so $\xi_c$ is always \textit{surjective}. Definitions~\ref{definition:unambiguous_representation} and~\ref{definition:minimized_representation} ensure that $\xi_c$ is well-defined and injective---hence bijective---on training data.

\begin{definition}[Unambiguous Representation]
\label{definition:unambiguous_representation}
Given $\mathcal{H} \cong_S \mathcal{Z}$, $\mathcal{H}$ has an \textit{unambiguous representation} w.r.t.\ $\mathcal{Z}$ at component $c \in \mathcal{C}$ if $\xi_c$ is well-defined, i.e., no two distinct reference values share the same hypothesis value on training data.
\end{definition}

Ambiguity means that one hypothesis value maps to two different reference values. The unambiguous condition rules this out.

\begin{definition}[Minimized Representation]
\label{definition:minimized_representation}
Given $\mathcal{H} \cong_S \mathcal{Z}$ and $\mathcal{H}$ has an unambiguous representation w.r.t.\ $\mathcal{Z}$ at $c \in \mathcal{C}$, $\mathcal{H}$ has a \textit{minimized representation} w.r.t.\ $\mathcal{Z}$ at $c$ if $|\operatorname{domain}(\xi_c)|$ is minimized, i.e.,
\[
|\operatorname{domain}(\xi_c)| = |\operatorname{image}(\xi_c)|.
\]
\end{definition}

\noindent\textbf{Justification.} Since $\xi_c$ is well-defined and surjective, by the pigeonhole principle, we always have $|\operatorname{domain}(\xi_c)| \geq |\operatorname{image}(\xi_c)|$. Thus, minimality holds exactly at equality.

\begin{definition}[Aligned Structure--Unambiguous Minimized Representation, AS-UMR]
\label{definition:aligned_structure-unambiguous_minimized_representation}
A model with hypothesis graph set $\mathcal{H}$ satisfies \textit{AS-UMR} if there exists a reference graph set $\mathcal{Z} \in \mathbb{Z}$ such that:
\begin{enumerate}
    \item $\mathcal{H} \cong_S \mathcal{Z}$ (structural alignment, Definition~\ref{definition:structure_alignment}); and
    \item for every $c \in \mathcal{C}$, $\mathcal{H}$ has, w.r.t.\ $\mathcal{Z}$ at $c$:
    \begin{enumerate}
        \item an unambiguous representation (Definition~\ref{definition:unambiguous_representation}), and
        \item a minimized representation (Definition~\ref{definition:minimized_representation}).
    \end{enumerate}
\end{enumerate}
\end{definition}

Intuitively: excessive compression introduces ambiguity and loses necessary information for compositional generalization, while insufficient compression (non-minimized representation) leaves spurious redundant information. The three conditions correspond to three design levers: structural alignment to architecture, unambiguous representation to training data diversity, and minimized representation to regularization.

\subsection{Necessity Assumptions}
We introduce two assumptions used only in the necessity direction. The sufficiency direction requires none of them.

\begin{assumption}[Seen Test Component Inputs]
\label{assumption:seen_test_component_inputs}
If a model provably enables compositional generalization (Definition~\ref{definition:compositional_generalization}), then all test component inputs have been seen in training:
\begin{align*}
    & \forall B \in \mathcal{D}_{\mathrm{test}}, \forall h^B \in H^B \setminus X^B, \\
    & \exists A \in \mathcal{D}_{\mathrm{train}}, \exists h^A \in H^A \setminus X^A: \mathbf{h}^A = \mathbf{h}^B,
\end{align*}
or an equivalent condition holds.
\end{assumption}

\paragraph{Justification.}
Because training and test distributions differ, a model that performs well on training data need not generalize. Without additional constraints (i.e., inductive bias), training cannot distinguish between compositional and non-compositional solutions. Consequently, a model cannot \textit{provably} generalize compositionally unless every component's test inputs have been observed during training (or an equivalent constraint is satisfied), which effectively decomposes the problem into in-domain subproblems. Moreover, as a heuristic argument, when correct test outputs are unseen, a trained model tends to reproduce only previously seen training outputs. Section~\ref{section:justification} provides further justification.

\paragraph{Remark.}
The qualifier ``provably'' in Theorem~\ref{theorem:necessary_and_sufficient_condition} is introduced by Assumption~\ref{assumption:seen_test_component_inputs}.

\begin{assumption}[Correct Training Predictions]
\label{assumption:correct_training_predictions}
The model correctly predicts all training samples:
\[
\forall A \in \mathcal{D}_{\mathrm{train}} : \hat{Y}^A = Y^A.
\]
\end{assumption}

This assumption prevents the invalid conclusion that a model satisfies AS-UMR simply because a violation of correct training predictions makes Definition~\ref{definition:compositional_generalization} vacuously true.

\section{Derivations}
We now prove Theorem~\ref{theorem:necessary_and_sufficient_condition}. Full proofs are deferred to Appendix~\ref{section:proofs}. The component-wise value mapping between hypothesis and reference nodes (Definition~\ref{definition:component-wise_value_mapping}) has a key property, captured by the following lemma (illustrated in Figure~\ref{fig:visual_proof_of_mappings_on_nodes}):

\begin{lemma}[Mappings on Nodes]
\label{lemma:mappings_on_nodes}
Let \(S_1\) and \(S_2\) be finite sets, with \(S_2\) fixed. A well-defined and surjective mapping \(f: S_1 \to S_2\) is injective if and only if \(|S_1|\) is minimized, i.e., \(|S_1| = |S_2|\).
\end{lemma}

\begin{figure}[htbp]
\centering
\begin{tikzpicture}[
    scale=0.475,
    set/.style={ellipse, draw, minimum width=1.5cm, minimum height=2cm},
    large_set/.style={ellipse, draw, minimum width=1.5cm, minimum height=2.5cm},
    element/.style={circle, draw=black, minimum size=0.1mm},
    set_name/.style={minimum size=5mm},
    arrow/.style={->, >=stealth', shorten >=1pt}
]

% Left side (Bijection case)
\node[set] (A1) at (0,0) {};
\node[set] (B1) at (4,0) {};
\node[set_name] at (-0.9,0) {$S_1$};
\node[set_name] at (4.8,0) {$S_2$};

\node[element] (a1) at (0,1) {};
\node[element] (a2) at (0,0) {};
\node[element] (a3) at (0,-1) {};

\node[element] (b1) at (4,1) {};
\node[element] (b2) at (4,0) {};
\node[element] (b3) at (4,-1) {};

\draw[arrow] (a1) -- (b1);
\draw[arrow] (a2) -- (b2);
\draw[arrow] (a3) -- (b3);

% Right side (Non-injective case)
\node[large_set] (A2) at (8,0) {};
\node[set] (B2) at (12,0) {};
\node[set_name] at (7.1,0) {$S_1'$};
\node[set_name] at (12.8,0) {$S_2$};

\node[element] (a4) at (8,1.5) {};
\node[element] (a5) at (8,0.5) {};
\node[element] (a6) at (8,-0.5) {};
\node[element] (a7) at (8,-1.5) {};

\node[element] (b4) at (12,1) {};
\node[element] (b5) at (12,0) {};
\node[element] (b6) at (12,-1) {};

\draw[arrow] (a4) -- (b4);
\draw[arrow] (a5) -- (b5);
\draw[arrow] (a6) -- (b6);
\draw[arrow] (a7) -- (b6);

\end{tikzpicture}
\caption{Visual proof of Lemma~\ref{lemma:mappings_on_nodes}: for a well-defined and surjective mapping, injectivity is equivalent to having the minimized domain size.
Left: when $|S_1|$ is minimized to equal $|S_2|$, the mapping is injective.
Right: when $|S_1'|$ is not minimized, multiple elements in $S_1'$ map to the same element in $S_2$, violating injectivity.}
\label{fig:visual_proof_of_mappings_on_nodes}
\end{figure}

\subsection{Necessity}
We first establish necessity. If compositional generalization holds, then by Assumption~\ref{assumption:correct_training_predictions} and Definition~\ref{definition:compositional_generalization}, all training and test graphs produce correct outputs. By Assumption~\ref{assumption:seen_test_component_inputs}, all test component inputs have been seen in training. Together, these imply that the hypothesis graph set $\mathcal{H}$ is itself a reference graph set (Definition~\ref{definition:reference_graph_set}).

Setting $\mathcal{Z} = \mathcal{H}$ immediately yields structural alignment (Definition~\ref{definition:structure_alignment}), since every node maps bijectively to itself. Unambiguous representation (Definition~\ref{definition:unambiguous_representation}) follows trivially. Minimized representation (Definition~\ref{definition:minimized_representation}) then follows from Lemma~\ref{lemma:mappings_on_nodes}. This yields the following result:

\begin{proposition}[Necessity]
\label{proposition:necessity}
A model provably enables compositional generalization (Definition~\ref{definition:compositional_generalization}) only if it has AS-UMR (Definition~\ref{definition:aligned_structure-unambiguous_minimized_representation}).
 
\end{proposition}

\subsection{Sufficiency}
We now prove sufficiency. Because each graph has a topological order, we proceed by (bottom-up) induction on the nodes of any test graph (see Appendix~\ref{sec:intuition_behind_the_inductive_step} for additional intuition).

The following lemma establishes the core inductive step. It requires: (1) structural alignment, (2) injectivity of $\xi_c$ for all $c \in \mathcal{C}$, following from unambiguous and minimized representations (Lemma~\ref{lemma:mappings_on_nodes}), and (3) the induction hypothesis that all input nodes of a given test non-input node match their training counterparts.

\begin{lemma}[Inductive Step]
\label{lemma:inductive_step}
For a hypothesis graph set $\mathcal{H}$, suppose the following two conditions hold:
\begin{align*}
    (1) \;\; & \exists \mathcal{Z} \in \mathbb{Z}: \mathcal{H} \cong_S \mathcal{Z}, \text{ and}\\
    (2) \;\; & \forall c \in \mathcal{C}, \xi_c \text{ is injective.}
\end{align*}
For all $B \in \mathcal{D}_{\mathrm{test}}, \forall z^B \in Z^B \setminus X^B$,
\begin{align*}
\text{if} \;\; (3) \;\;
    & \forall i \in \{1, \dots, n\}, \\
    & \exists C_i \in \mathcal{D}_{\mathrm{train}}, \exists z^{C_i} \in Z^{C_i}: \\
    & z^{C_i} = z^B_i, h^{C_i} = h^B_i, \\
\text{then} \quad\quad\;\;
    & \exists A \in \mathcal{D}_{\mathrm{train}}, \exists z^A \in Z^A \setminus X^A: \\
    & z^A = z^B, h^A = h^B.
\end{align*}

\end{lemma}

Figure~\ref{fig:illustrative_proof_of_the_inductive_step} provides a diagrammatic proof sketch.

\begin{figure}[!ht]
\centering
\begin{tikzpicture}[
    node/.style={draw, circle, minimum size=8mm},
    train/.style={node},
    test/.style={node, fill=gray!20},
    precondition/.style={thick},
    assigned/.style={thick, dashed}
]

% Training nodes (left cluster)
\node[train] (zA1) {$z^A_i$};
\node[train, below right=1.0cm and 0.5cm of zA1] (zA2) {$z^{C_i}$};
\node[test, right=1.3cm of zA1] (hA1) {$z^B_i$};

% Test nodes (right cluster)
\node[train, right=3cm of zA1] (zB1) {$h^A_i$};
\node[train, below right=1.0cm and 0.5cm of zB1] (zB2) {$h^{C_i}$};
\node[test, right=1.3cm of zB1] (hB1) {$h^B_i$};

\draw[precondition] (zA1) -- node[above] {$\alpha$} (hA1);
\draw[precondition] (zA2) -- node[right] {$\beta_z$} (hA1);
\draw[precondition] (zB2) -- node[right] {$\beta_h$} (hB1);

\draw[assigned] (zA2) -- node[left] {I} (zA1);
\draw[assigned] (zB2) -- node[left] {II} (zB1);
\draw[assigned] (hB1) -- node[above] {III}(zB1);

\end{tikzpicture}

\caption{
Illustrative proof of the inductive step (Lemma~\ref{lemma:inductive_step}). At each step, for a test node $z^B$, there exists a training node $z^A$ with $\mathbf{z}^A = \mathbf{z}^B$ (Definition~\ref{definition:reference_graph_set}). For each input index $i$, there is a training node $z^{C_i}$. This figure shows the relationships among the nodes. Training nodes are white; test nodes are gray. Two nodes connected by any style of line have the same value. Solid lines represent preconditions. $\alpha$ follows from equal reference inputs. $\beta_z$ and $\beta_h$ follow from the induction hypothesis. Dashed lines are assigned step by step. I is derived from $\alpha$ and $\beta_z$. II follows from I and the injective mapping in training, i.e., condition (2). III follows from II and $\beta_h$. This applies to all input nodes $i$, thus $\mathbf{h}^A=\mathbf{h}^B$. By the deterministic property of components (Definition~\ref{definition:component}), $\mathbf{z}^A=\mathbf{z}^B$ implies $z^A = z^B$ and $\mathbf{h}^A=\mathbf{h}^B$ implies $h^A = h^B$.
}
\label{fig:illustrative_proof_of_the_inductive_step}
\end{figure}

For any test sample $B$, consider one of its \textbf{output nodes}, say $z^B$. By applying Lemma~\ref{lemma:inductive_step} inductively, we obtain a training sample $A$ such that $z^A = z^B$ and $h^A = h^B$. By the premise of Definition~\ref{definition:compositional_generalization}, training predictions are correct, and thus we have $h^A = z^A$. Combining these equalities gives $h^B = h^A = z^A = z^B$, thus $h^B = z^B$. Since $z^B$ is an arbitrary output node, we have $\hat{Y}^B = Y^B$. This reasoning, applied to every test sample, establishes the following result:

\begin{proposition}[Sufficiency]
\label{proposition:sufficiency}
A model provably enables compositional generalization (Definition~\ref{definition:compositional_generalization}) if it has AS-UMR (Definition~\ref{definition:aligned_structure-unambiguous_minimized_representation}).

\end{proposition}

\section{Example Approach}
\label{section:example_approach}
We now present an example approach that leverages AS-UMR (Definition~\ref{definition:aligned_structure-unambiguous_minimized_representation}) to guide architecture, training data, and regularization design.

\subsection{Approach}
\label{sec:approach}
We use entropy minimization to enable the minimized representation. For a component \(c\in\mathcal{C}\), its entropy is the entropy of the training distribution over its output node.

\begin{definition}[Component-wise Minimum Entropy]
\label{definition:component-wise_minimum_entropy}
A non-output component \(c\in\mathcal{C}\) has \textit{minimum entropy} if its entropy cannot be reduced by modifying \(c\) alone while keeping all other components fixed, without increasing the training loss. A model has \textit{component-wise minimum entropy} if every non-output component \(c\in\mathcal{C}\) has minimum entropy.
\end{definition}

\begin{lemma}[Minimum Entropy Implies Minimized Representation]
\label{lemma:minimum_entropy_implies_minimized_representation}
Given that $\mathcal{H} \cong_S \mathcal{Z}$ (Definition~\ref{definition:structure_alignment}) and $\mathcal{H}$ has an unambiguous representation (Definition~\ref{definition:unambiguous_representation}) w.r.t.\ $\mathcal{Z}$ at $c \in \mathcal{C}$, if $c$ has minimum entropy (Definition~\ref{definition:component-wise_minimum_entropy}), then $\mathcal{H}$ has a minimized representation (Definition~\ref{definition:minimized_representation}) w.r.t.\ $\mathcal{Z}$ at $c$.
\end{lemma}
The three conditions of AS-UMR correspond to three design levers:
\begin{enumerate}
    \item \textbf{Structural alignment} (Definition~\ref{definition:structure_alignment}) is achieved by designing components and their composition based on prior knowledge of the task's compositional structure---for example, using modular networks that explicitly mirror the hierarchical composition of the input. In particular, if all samples share the same structure, this step reduces to static architecture design. The necessity of such prior knowledge is further discussed in Section~\ref{section:the_necessity_of_prior_knowledge_for_structural_alignment}.
    \item \textbf{Unambiguous representation} (Definition~\ref{definition:unambiguous_representation}) is achieved by preparing training data so that, if Algorithm~\ref{algorithm:entropy_minimization_regularization} (Section~\ref{sec:entropy_minimization_regularization}) reaches:
    \begin{enumerate}[label=(\alph*)]
        \item correct training predictions, and
        \item component-wise minimum entropy (Definition~\ref{definition:component-wise_minimum_entropy}),
    \end{enumerate}
    then there exists $\mathcal{Z} \in \mathbb{Z}$ such that every $c \in \mathcal{C}$ has an unambiguous representation w.r.t.\ $\mathcal{Z}$.
    \item \textbf{Minimized representation} (Definition~\ref{definition:minimized_representation}) is achieved by minimizing the entropy of each component's output (Lemma~\ref{lemma:minimum_entropy_implies_minimized_representation}); this is enabled by training with entropy minimization regularization (Algorithm~\ref{algorithm:entropy_minimization_regularization}), which injects Gaussian noise and applies a power constraint to remove redundant information.
\end{enumerate}
The resulting procedure is summarized in Algorithm~\ref{algorithm:example_approach}.

\begin{algorithm}[!ht]
\caption{Example approach}
\label{algorithm:example_approach}
\begin{algorithmic}[1]
\State Use prior knowledge to design components and their composition so that the resulting computational graph is structurally aligned with the task's compositional hierarchy for every potential sample.
\State \label{step:training_data}Prepare training data so that, if Algorithm~\ref{algorithm:entropy_minimization_regularization} reaches:
\begin{enumerate}[label=(\alph*)]
    \item correct training predictions, and
    \item component-wise minimum entropy (Definition~\ref{definition:component-wise_minimum_entropy}),
\end{enumerate}
then there exists $\mathcal{Z} \in \mathbb{Z}$ such that every $c \in \mathcal{C}$ has an unambiguous representation w.r.t.\ $\mathcal{Z}$.
\State Train with entropy minimization regularization (Algorithm~\ref{algorithm:entropy_minimization_regularization}) on the output nodes of all non-output components.
\end{algorithmic}
\end{algorithm}

\begin{corollary}[Example Approach Verification]
\label{corollary:example_approach_verification}
If component-wise minimum entropy (Definition~\ref{definition:component-wise_minimum_entropy}) is achieved, a model trained according to Algorithm~\ref{algorithm:example_approach} enables compositional generalization (Definition~\ref{definition:compositional_generalization}).
\end{corollary}

Section~\ref{sec:minimal_example} and Appendix~\ref{sec:scan_jump_task} provide concrete instantiations. In these examples, the training data is fixed; hence, Step~\ref{step:training_data} simplifies to verifying the condition.

\subsection{Entropy Minimization Regularization}
\label{sec:entropy_minimization_regularization}
We use regularization to achieve component-wise minimum entropy (Definition~\ref{definition:component-wise_minimum_entropy}). We adopt a particularly efficient entropy-reduction mechanism: the Gaussian channel with a power constraint \citep[see, e.g.,][p.~261]{cover2012elements}.

\begin{definition}[Gaussian Channel with Power Constraint $P$]
\label{definition:gaussian_channel_with_power_constraint_p}
\begin{align*}
    & Y = X + Z, \quad Z \sim \mathcal{N}(0, N),
    && \frac{1}{n}\sum^n_{i=1}x^2_i \leq P.
\end{align*}
\end{definition}

Here, $X$ is the channel input and $Y$ is the channel output. $n$ is the dimension of $X$. Building on this, we define a regularization algorithm (Algorithm~\ref{algorithm:entropy_minimization_regularization}) that adds noise to the output of each non-output component and enforces the power constraint with an activity regularization term, as in \citet{li2019compositional}.

\begin{algorithm}[!ht]
\caption{Entropy minimization regularization}
\label{algorithm:entropy_minimization_regularization}
\begin{algorithmic}[1]
\Statex For each non-output component output node $h$, replace $h$ with $h'$ and the training loss $\mathcal{L}$ with $\mathcal{L}'$:
\begin{align*}
    & h' = h + \mathcal{N}(0, \alpha),
    && \mathcal{L}' = \mathcal{L} + \beta \|h\|^2.
\end{align*}
The hyperparameters $\alpha$ and $\beta$ control the noise variance and the regularization strength, respectively.
\end{algorithmic}
\end{algorithm}

\paragraph{Information-theoretic intuition.}
The penalty term \(\beta\|h\|^2\) in Algorithm~\ref{algorithm:entropy_minimization_regularization} compresses \(\|h\|^2\) for every training sample, which effectively amounts to compressing \(P\) towards zero. Consequently, the Gaussian channel capacity \(C = \frac{1}{2}\log_2(1 + P/\alpha)\) tends to zero. For any required information rate \(R\), the actual mutual information \(I(h;h')\) satisfies \(0 \le R \le I(h;h') \le C\). In the limit of convergence, these inequalities become tight: \(R = I(h;h') = C\), indicating that \(h\) is compressed to have the minimum entropy for the required rate.

This corresponds to the notion of component-wise minimum entropy (Definition~\ref{definition:component-wise_minimum_entropy}): each component's output has been compressed to the point where any further reduction would violate the information requirement \(R\), which in turn would increase the training loss.

\paragraph{Remarks.}
When a component contains multiple layers, this regularization can be applied at every layer. Also, the entropy minimization regularization is distinct from the entropy regularization commonly used in reinforcement learning, which \textit{increases} entropy to encourage exploration.

\section{Minimal Example}
\label{sec:minimal_example}
To illustrate the concepts and their application, we present a minimal worked example. An extension to the more complex SCAN jump task appears in Appendix~\ref{sec:scan_jump_task}.

\subsection{Task}
\label{section:task}
The task has three binary inputs $x_1, x_2, x_3$ and one binary output $y$ (three input nodes, one output node). We use the exclusive-OR (XOR) operation, denoted $\oplus$. The ground-truth function is given in Algorithm~\ref{algorithm:ground-truth_function} (other ground-truth decompositions are possible), and the train/test split appears in Table~\ref{tab:values_for_the_example}. Note that the intermediate value $z$ is not given as part of the data.

\begin{algorithm}[!ht]
\caption{Ground-truth function}
\label{algorithm:ground-truth_function}
\begin{algorithmic}[1]
\Statex Given an input $[x_1, x_2, x_3] \in \{0,1\}^3$, the output $y$ is computed as:
\State $z = x_1 \oplus x_2 \in \{0,1\}$
\State $y = z \oplus x_3 \in \{0,1\}$
\end{algorithmic}
\end{algorithm}

\begin{table}[!ht]
    \caption{Values for the minimal example. The upper block (a--f) is training data; the lower block (g, h) is test data. The intermediate value $z$ is not given as part of the data.}
    \label{tab:values_for_the_example}
    \centering
    \begin{tabular}{c|ccc|c|c}
         & $x_1$ & $x_2$ & $x_3$ & $z$ & $y$  \\
         \hline
         a & 0 & 0 & 0 & 0 & 0 \\
         b & 0 & 1 & 0 & 1 & 1 \\
         c & 1 & 0 & 1 & 1 & 0 \\
         d & 1 & 1 & 1 & 0 & 1 \\
         e & 0 & 0 & 1 & 0 & 1 \\
         f & 0 & 1 & 1 & 1 & 0 \\
         \hline
         g & 1 & 0 & 0 & 1 & 1 \\
         h & 1 & 1 & 0 & 0 & 0
    \end{tabular}
\end{table}

Algorithm~\ref{algorithm:ground-truth_function} induces a reference graph set (Definition~\ref{definition:reference_graph_set}) for the dataset in Table~\ref{tab:values_for_the_example}: predictions are correct on all samples, and the test inputs to each component---$(x_1,x_2)$ and $(z,x_3)$---both appear in training. Any full test input triple $(x_1,x_2,x_3)$ is unseen during training; hence, compositional generalization is required. The model must learn how $x_1$ and $x_2$ interact to produce $z$, and how $z$ and $x_3$ interact to produce $y$, and then recombine these subskills on novel input combinations.

\subsection{Algorithm}
\label{section:xor_algorithm}
We instantiate Algorithm~\ref{algorithm:example_approach}.

\paragraph{Structural Alignment.}
We construct a model (Algorithm~\ref{algorithm:forward_pass_for_the_model}). To enable correct training predictions, the embedding component $E$ must be bijective (0 and 1 have distinct embeddings; otherwise, training inputs are indistinguishable), so we do not consider it in the analysis. The hypothesis graphs from Algorithm~\ref{algorithm:forward_pass_for_the_model} and the reference graphs from Algorithm~\ref{algorithm:ground-truth_function} have structural alignment (Definition~\ref{definition:structure_alignment}).

\begin{algorithm}[!ht]
\caption{Forward pass of the model}
\label{algorithm:forward_pass_for_the_model}
\begin{algorithmic}[1]
\Statex Let $E \in \mathbb{R}^{2 \times m}$ be an embedding matrix. $f_h$ and $f_{\hat{y}}$ denote feed-forward networks. Given a one-hot input $X \in \{0, 1\}^{3 \times 2}$:
\State $e = XE \in \mathbb{R}^{3 \times m}$
\State $h = f_h(\text{concat}(e_1^T, e_2^T)) \in \mathbb{R}^m$
\State $\hat{y} = \text{softmax}(f_{\hat{y}}(\text{concat}(h, e_3^T))) \in \mathbb{R}^2$
\end{algorithmic}
\end{algorithm}

\paragraph{Unambiguous Representation.}
The model also satisfies the unambiguous representation condition (Definition~\ref{definition:unambiguous_representation}).

\begin{lemma}[Unambiguous Representation Verification]
\label{lemma:unambiguous_representation_verification}
Given correct training predictions, the training samples shown in Table~\ref{tab:values_for_the_example} have an unambiguous representation (Definition~\ref{definition:unambiguous_representation}) for every non-output component.
\end{lemma}

\begin{proof}[Proof sketch]
The architecture is shared across all samples, and the hidden node $h$ is the output of the only non-output component. By inspecting the training samples, we find that whenever two training samples can produce the same hypothesis value $h$, they also share the same reference value $z$. Hence, the representation is unambiguous.
\end{proof}

\paragraph{Minimized Representation.}
To satisfy the minimized representation condition (Definition~\ref{definition:minimized_representation}), we apply entropy minimization regularization (Algorithm~\ref{algorithm:entropy_minimization_regularization}) to each hidden layer of $f_h$. This drives the number of distinct training outputs of the component to its minimum, i.e., the number of possible reference values.

\begin{corollary}[Minimal Example Verification]
\label{corollary:minimal_example_verification}
In the minimal example task (Section~\ref{section:task}), if component-wise minimum entropy (Definition~\ref{definition:component-wise_minimum_entropy}) is achieved, the algorithm designed in Section~\ref{section:xor_algorithm} enables compositional generalization (Definition~\ref{definition:compositional_generalization}).
\end{corollary}

\subsection{Results}
Experimental details are given in Appendix~\ref{section:experiment_details}; code is available in the repository. As a baseline, we use a fully connected network that takes the concatenated embeddings of $x_1, x_2, x_3$ as a single input (flattened), demonstrating that the task is non-trivial. We report the mean and standard deviation of accuracy over 25 runs.

The upper block of Table~\ref{tab:experiment_results_in_accuracy} shows that the full model (which satisfies all three conditions) achieves perfect test accuracy, while the baseline fails. Neural networks generally do not learn XOR-like functions in an out-of-distribution setting, so the baseline (and the ablation models below) perform below chance, producing a clear contrast.

\begin{table}[!ht]
    \caption{Experimental results in accuracy (mean $\pm$ std).}
    \label{tab:experiment_results_in_accuracy}
    \centering
    \begin{tabular}{lrr}
        & \multicolumn{1}{c}{Train} & \multicolumn{1}{c}{Test} \\
        \hline
        \multicolumn{3}{c}{Main results} \\
        Baseline & 1.0 $\pm$ 0.0 & 0.0 $\pm$ 0.0 \\
        Full model & 1.0 $\pm$ 0.0 & 1.0 $\pm$ 0.0 \\
        \hline
        \multicolumn{3}{c}{Ablation study} \\
        No structural alignment & 1.0 $\pm$ 0.0 & 0.0 $\pm$ 0.0 \\
        Broken unambiguity & 1.0 $\pm$ 0.0 & 0.0 $\pm$ 0.0 \\
        No entropy minimization & 1.0 $\pm$ 0.0 & 0.0 $\pm$ 0.0
    \end{tabular}
\end{table}

\subsection{Ablation Study}
We perform an ablation study to show that removing any single condition undermines the compositional generalization guarantee:
\begin{itemize}
    \item \textit{No structural alignment}: use the baseline architecture, breaking structural alignment.
    \item \textit{Broken unambiguity}: remove the last two training samples (e and f) from Table~\ref{tab:values_for_the_example}, breaking the unambiguous representation at the hidden node (see below).
    \item \textit{No entropy minimization}: remove the entropy minimization regularization, breaking the minimized representation.
\end{itemize}

The lower block of Table~\ref{tab:experiment_results_in_accuracy} shows that none of the ablation models generalize to the test set, consistent with our theoretical analysis that each condition is individually necessary for the generalization guarantee.

To illustrate the violation of the unambiguous representation concretely, suppose samples e and f are removed, and the hidden representation has $h = x_2$ on the remaining training samples. The pairs $(h, x_3)$ are all distinct across training samples, so the model can still produce correct training outputs. The corresponding $\mathcal{Z}$ is still a reference graph set (Definition~\ref{definition:reference_graph_set}) because the test inputs to both components, $(x_1, x_2)$ and $(z, x_3)$, appear in training. However, the pair of samples a and c has $h^a = h^c$ but $z^a \neq z^c$, violating the unambiguous representation and thus invalidating the generalization guarantee.

\section{Discussion}
Additional discussion is deferred to Appendix~\ref{section:additional_discussion}.

\subsection{Attention Mechanism}
\label{sec:attention_mechanism}
The attention mechanism is widely used in modern neural networks. We analyze how it increases the likelihood that test component inputs have been seen during training, a prerequisite for compositional generalization (Assumption~\ref{assumption:seen_test_component_inputs}). We introduce:

\begin{definition}[Effectively Equal Component Inputs]
Two inputs $\mathbf{h}^A, \mathbf{h}^B$ for the same component are \textit{effectively equal} if the component implements a commutative operation and the inputs are equal up to permutation of their entries.
\end{definition}

A commutative operation is invariant to input ordering (e.g., addition). Two effectively equal component inputs can be treated as equal ($\mathbf{h}^A = \mathbf{h}^B$) because permuting the input nodes before feeding them into the component leaves the output unchanged. Thus, the chance of unseen component inputs is reduced.

The attention mechanism is naturally compatible with effectively equal component inputs due to its commutative summation operation. Attention combines an attention weight vector $\mathbf{a} \in \mathbb{R}^k$ with a value matrix $V \in \mathbb{R}^{d \times k}$ to produce a context vector:
\[
\text{context vector} = V\mathbf{a} = \sum^k_{i=1} a_i\mathbf{v}_i \in \mathbb{R}^d,
\]
where $k$ is the dimension of the attention weight vector, and $d$ is the dimension of each value vector.

Furthermore, the attention mechanism has other properties that help reduce unseen component inputs: attention typically focuses on only a sparse subset of inputs; the query at each position is computed solely from the layer input at that position, and in self-attention, keys also share this locality.

\subsection{Seen Test Component Inputs}
We further justify the seen test component inputs assumption (Assumption~\ref{assumption:seen_test_component_inputs}).

\subsubsection{Justification}
\label{section:justification}
We consider two cases based on whether the correct test outputs lie within the support of the training output distribution:
\begin{itemize}
    \item \textbf{General case}: Test inputs are unseen in training (correct test outputs may or may not have been observed).
    \item \textbf{Special case}: Correct test outputs are unseen in training (a subcase of the general case).
\end{itemize}
The general-case argument alone can support the assumption; the special-case argument provides further justification.

\paragraph{General case.}
Because the training and test distributions differ, the set of solutions that fit the training data may include both compositional and non-compositional models. Training loss alone cannot provably distinguish between them.

Additional information (i.e., inductive bias) is required to favor compositional solutions. This is effectively equivalent to ensuring that every component's inputs have been seen in training---for instance, via architecture design (e.g., attention mechanisms in Section~\ref{sec:attention_mechanism} or linear models in Figure~\ref{figure:illustrative_examples}, left), data augmentation, regularization, or pretraining. The probability of converging to a correct compositional solution drops sharply as the number of unseen combinations grows.

\begin{figure}[!ht]
\centering
\begin{tikzpicture}[scale=1]
\draw[thick,dashed] (2,0.5) -- (2,2.5) node[right] {$M_2$};
\draw[thick] (2.1,0.5) -- (2.1,1.5) -- (3, 1.8) node[above] {$M_1$};
\fill[black] (1,1) circle (0.1);
\fill[black] (1,2) circle (0.1);
\draw[black] (3,1) circle (0.1);
\end{tikzpicture}
\quad\quad\quad\quad
\begin{tikzpicture}[scale=1]
\draw[thick] (0.9,1.5) -- (2,1.5) -- (3, 2);
\draw[thick] (2,0.5) -- (2,1.4) -- (3, 1.9);
\fill[black] (1,1) circle (0.1);
\fill[black] (0.9,1.9) rectangle (1.1, 2.1);
\draw[black] (3,1) circle (0.1);
\end{tikzpicture}
\caption{Left (general case): insufficient information to determine whether model $M_1$ or $M_2$ is the correct compositional solution. Right (special case): a model with two binary output nodes (corresponding to two decision boundaries) reproduces only the training output combinations, mapping test inputs to seen outputs and thus producing incorrect test predictions.}
\label{figure:illustrative_examples}
\end{figure}

\paragraph{Special case.}
As a heuristic argument, consider the full dataset. We say two samples merge if their representations coincide at some layer. 

Due to the determinism of neural networks at inference time, once two samples merge, they stay merged in the subsequent layers and thus have the same final output. Hence, a test sample has a correct prediction only if its representations are distinct from those of any training sample in \textit{all} layers. This is a strong requirement, as it must hold for every layer.
Also, since test predictions do not contribute to the training loss, there is no direct mechanism to avoid the spurious merge. Moreover, when incorrectly merged training samples are separated to reduce training loss, some of them may merge with test samples that were previously not merged with them.

As a result, test inputs tend to map to training outputs, yielding incorrect test predictions under correct training predictions. Figure~\ref{figure:illustrative_examples} (right) illustrates this effect.

\subsubsection{On the Negation of ``Provably Enables''}
The logical negation of ``provably enables compositional generalization'' is ``does not provably enable'', not ``provably does not enable''. This distinction is important for three reasons.

\paragraph{(A) ``Does Not Provably Enable'' Matches the Research Paradigm.}
Compositional generalization by definition avoids enumerating all possible component combinations in training---the number grows exponentially and cannot be exhausted. Similarly, testing every possible combination to rule out all generalization failures is also infeasible due to this exponential growth. Even if a model makes correct predictions on a subset of the test combinations, this does not imply that it will generalize to other unseen combinations. Thus, only theoretical guarantees are reliable in such cases.

When a model is not provably compositional, there is no principled reason to trust its performance on unseen combinations. This caution is warranted on its own; we need not prove the much stronger claim that the model \textit{provably does not} generalize.

\paragraph{(B) ``Provably Does Not Enable'' Has a Counterexample.}
Guaranteed failure cannot be proven because a trivial counterexample always exists: a model initialized with parameters that produce the correct output for every sample, without training. While practically improbable, this mathematical possibility invalidates any universal claim of guaranteed failure.

\paragraph{(C) Tendency Toward ``Provably Does Not Enable''.}
Although guaranteed failure cannot be proven, the analysis in Section~\ref{section:justification} suggests a tendency toward it:
\begin{itemize}
    \item In the general case, training cannot distinguish compositional from non-compositional solutions; the probability of converging to a correct solution decays rapidly as the number of unseen combinations grows.
    \item In the special case, neural networks are intrinsically biased toward mapping unseen test inputs to outputs observed during training.
\end{itemize}

\subsection{Alternative Definition}
\label{sec:alternative_definition}
In some contexts, compositional generalization is defined as requiring all test component inputs to have been observed in training.

\begin{definition}[Alternative Compositional Generalization]
\label{definition:alternative_compositional_generalization}
Alternative compositional generalization is compositional generalization (Definition~\ref{definition:compositional_generalization}) with the additional requirement that all test component inputs are seen in training:
\begin{align*}
    (1) \;\; & \text{If} && \forall A \in \mathcal{D}_{\mathrm{train}}: \hat{Y}^A = Y^A, \\
    & \text{then} && \forall B \in \mathcal{D}_{\mathrm{test}}: \hat{Y}^B = Y^B, \text{ and} \\
    (2) \;\; & && \forall B \in \mathcal{D}_{\mathrm{test}}, \forall h^B \in H^B \setminus X^B, \\
    & && \exists A \in \mathcal{D}_{\mathrm{train}}, \exists h^A \in H^A \setminus X^A: \mathbf{h}^A = \mathbf{h}^B.
\end{align*}
\end{definition}

Under this definition, Assumption~\ref{assumption:seen_test_component_inputs} (the necessity of seen test component inputs) becomes an integral part of the definition. Consequently, Theorem~\ref{theorem:necessary_and_sufficient_condition} is adapted to Theorem~\ref{theorem:alternative_necessary_and_sufficient_condition}, which no longer relies on Assumption~\ref{assumption:seen_test_component_inputs}; hence, the qualifier ``provably'' is removed.

\begin{theorem}[Alternative Necessary and Sufficient Condition]
\label{theorem:alternative_necessary_and_sufficient_condition}
A model enables alternative compositional generalization (Definition~\ref{definition:alternative_compositional_generalization}) if and only if it has AS-UMR (Definition~\ref{definition:aligned_structure-unambiguous_minimized_representation}).

\end{theorem}

\subsection{Prior Knowledge and Empirical Validation}
\subsubsection{The Necessity of Prior Knowledge for Structural Alignment}
\label{section:the_necessity_of_prior_knowledge_for_structural_alignment}
One might worry that requiring prior knowledge of the true compositional hierarchy (Definition~\ref{definition:structure_alignment}) renders the theory impractical. This concern, however, reflects a necessity rather than a limitation. Proposition~\ref{proposition:necessity} shows that structural alignment is not an arbitrary design choice but is inherent to the very nature of compositional generalization. Making this requirement explicit thus clarifies what prior knowledge is indispensable and provides a concrete target for architecture design and meta-learning.

As noted in Remark~\ref{remark:structure_as_a_program}, the required structure need not be sample-specific but can be a single uniform program that applies to all instances of the task.

\subsubsection{Empirical Validation}
We do not include large-scale empirical evaluations on real-world open-domain benchmarks for two complementary reasons. First, satisfying all three conditions---especially structural alignment---in these tasks is inherently difficult, as it requires acquiring reliable prior knowledge about complex compositional hierarchies. Second, our contributions are theoretical: the results are proved mathematically. Using synthetic and controlled datasets is standard practice in theoretical studies of compositional generalization \citep[see, e.g.,][]{jarvis2023on,NEURIPS2023_15f6a108,ahuja2024on,lippl2025when}. The absence of these large-scale experiments does not weaken our conclusions; it reflects the intrinsic difficulty of compositional generalization, which our theory precisely characterizes.

\subsection{Future Work}
Our theoretical results identify structural knowledge as the central bottleneck for compositional generalization: structure and training data are two complementary dimensions of information. Decomposing a problem into compositional parts typically requires domain-specific prior knowledge. A potential direction is to systematically harvest structural patterns from disciplines such as linguistics, cognitive science, and biology.

With such knowledge, the ``structure-driven automated learning'' paradigm could offer a pathway to translating our theoretical conditions into scalable and practical solutions. Constructing an explicit function-interface call graph for structural alignment, combined with diverse end-to-end training data for unambiguous representation and appropriate regularization for minimized representation, would enable automated learning of a neural network. This does not require manually specified internal details such as intermediate outputs.

\section{Related Work}
\label{sec:related_work}

\subsection*{Compositional Generalization and Deep Learning}
Compositional generalization \citep{fodor1988connectionism,lake2018generalization} is the defining challenge when test samples lie outside the training distribution. Recent work has sought general inductive biases for this capability \citep{goyal2022inductive}, including the Consciousness Prior framework \citep{bengio2017consciousness,butlin2023consciousness}.

A closely related field is causal learning, rooted in classical AI \citep{pearl2003causality} and developed through do-calculus \citep{pearl2009causality} and interventional analysis \citep{ahuja2023interventional}. Causality gives rise to the principle of Independent Causal Mechanisms (ICMs) \citep{scholkopf2021toward}. Component recombination in our setting corresponds to the counterfactual scenario where the joint input distribution is intervened upon to produce novel combinations with zero probability under the training distribution---a form of covariate shift.

Connectionist models with distributed representations describe objects through sets of factors. While such representations can, in principle, combine to describe unseen objects \citep{hinton1990mapping}, it has been argued that they do not by themselves solve general compositional generalization \citep{fodor1988connectionism,marcus1998rethinking,mittal2022modular,dziri2023faith,jiang-etal-2024-peek,mirzadeh2025gsmsymbolic}. Deep learning models are modern instances of parallel distributed processing (PDP) systems, and their empirical successes \citep{openai2023gpt4,DeepSeekR1} motivate equipping them with principled compositional generalization guarantees.

\subsection*{Recent Theoretical Work}
Several recent theoretical studies examine compositional generalization from complementary angles. \citet{jarvis2023on} show that modular architectures alone do not guarantee compositional generalization without aligned dataset structure, emphasizing the role of training dynamics and low-rank substructures. \citet{lippl2025when} develop a kernel-theoretic analysis revealing that compositional models are limited to ``conjunction-wise additive'' computations, which rules out transitive generalization. \citet{NEURIPS2023_15f6a108} derive conditions on data-generating processes and model architectures via an identifiable representation framework, proving that generalization requires sufficient latent support and compositional function structure. \citet{ahuja2024on} establish provable compositional generalization guarantees for sequence-to-sequence models, showing that limited-capacity architectures generalize when training distributions are sufficiently diverse. \citet{fu2025on} propose an invariant measure of cross-task learnability and highlight the roles of data structure and algorithm--rule compatibility. \citet{mahajan2025compositional} introduce Compositional Risk Minimization for compositional shift, achieving provable extrapolation to novel attribute combinations within the discrete affine hull of the training support, using flexible additive energy models. \citet{ijcai2024p0533} propose a neuro-symbolic formalism that defines compositional complexity via computational DAGs and locus-of-influence metrics, analyzing how architectures such as CNNs and Transformers encode hierarchical structure. More detailed comparisons are in Appendix~\ref{section:comparison_with_related_work}. Inspired by these studies, we provide a necessary and sufficient condition that unifies architecture, training data, and regularization.

\subsection*{Recent Algorithmic Approaches}
Beyond architecture design \citep{andreas2016neural,russin2019compositional,soulos2024compositional} and data augmentation \citep{akyurek2023lexsym}, three main lines of work target compositional generalization: disentangled representation learning, attention mechanisms, and meta-learning.

\textit{Disentangled representations} \citep{brady2025interaction,NEURIPS2022_9f9ecbf4,wiedemer2024provable} are typically learned in an unsupervised fashion and used as feature extractors for downstream recombination. Early methods exploited statistical independence \citep{higgins2017beta}; later work extended the notion with symmetry transformations \citep{higgins2018towards}, leading to symmetry-based disentanglement \citep{NEURIPS2020_9a02387b}. Representational compositionality has also been characterized through algorithmic information theory \citep{elmoznino2025towards}.

\textit{Attention mechanisms} \citep{vaishnav2023gamr} are central to modern deep learning. Transformers \citep{vaswani2017attention,shi2024exedec,schug2025attention} are built on self-attention. Recurrent Independent Mechanisms \citep{goyal2021recurrent} use attention together with node names for variable binding. The Global Workspace architecture \citep{goyal2021coordination} extends this idea with limited-capacity global communication to enable knowledge interchangeability, and the discrete-valued communication bottleneck \citep{liu2021discrete} further improves generalization.

\textit{Meta-learning} \citep{lake2023human,pmlr-v202-wu23d,schug2024discovering} trains a meta-learner on a distribution of tasks and applies it to a target task---each task has its own compositional training and test data. When ICMs are available, they can be used to generate meta-training tasks \citep{scholkopf2021toward}. Meta-reinforcement learning has also been applied to causal reasoning \citep{dasgupta2019causal}, and meta-learning can capture the adaptation speed to discover causal structure \citep{bengio2020a,lippe2022efficient}.

\section{Conclusion}

We derived a necessary and sufficient condition for provable compositional generalization in neural networks, supported by rigorous mathematical proofs and formal verification in Lean~4. 
The condition yields a unified inductive bias that jointly governs architectural design, training data properties, and regularization strategies. Building on the condition, we developed an example algorithmic approach, illustrated it through a controlled minimal example, and further demonstrated the condition on the SCAN jump task. Our work provides a theoretical characterization of provable compositional generalization.

\bibliography{aaai21}

\appendix

\onecolumn
\section*{Appendix Contents}
\begin{center}
\begin{tabular}{llr}
    & Contents & Page \\
    \hline
    Appendix~\ref{section:list_of_symbols} & List of Symbols & \pageref{section:list_of_symbols} \\
    Appendix~\ref{section:proofs} & Proofs & \pageref{section:proofs} \\
    Appendix~\ref{section:experiment_details} & Experiment Details & \pageref{section:experiment_details} \\
    Appendix~\ref{section:additional_discussion} & Additional Discussion & \pageref{section:additional_discussion} \\
    Appendix~\ref{section:comparison_with_related_work} & Comparison with Related Work & \pageref{section:comparison_with_related_work} \\
    Appendix~\ref{sec:scan_jump_task} & SCAN Jump Task & \pageref{sec:scan_jump_task} \\
    Appendix~\ref{section:cogs_task} & COGS Task & \pageref{section:cogs_task}
\end{tabular}
\end{center}

\section{List of Symbols}
\label{section:list_of_symbols}
\renewcommand{\arraystretch}{1.02} % Adjust row height for readability
\begin{center}
\begin{tabularx}{0.74\linewidth}{l X}
\textbf{Symbol} & \textbf{Description} \\
\hline
\multicolumn{2}{c}{\textit{Datasets and Samples} (First Appearance: Section~\ref{section:definitions_and_assumptions})} \\
$\mathcal{D}_{\mathrm{train}}$ & Training dataset \\
$\mathcal{D}_{\mathrm{test}}$ & Test dataset \\
$\mathcal{D}$ & Full dataset, $\mathcal{D} = \mathcal{D}_{\mathrm{train}} \cup \mathcal{D}_{\mathrm{test}}$ \\
$X$ & Input of a sample \\
$Y$ & Ground-truth output of a sample \\
$\hat{Y}$ & Model-predicted output of a sample \\
$A, C, \dots$ & Indices for training samples \\
$B, D, \dots$ & Indices for test samples \\
\hline
\multicolumn{2}{c}{\textit{Computational Graphs and Nodes} (Section~\ref{section:definitions_and_assumptions})} \\
$H$ & A hypothesis graph (generated by a model) or the set of nodes in it \\
$H \setminus X$ & Set of non-input nodes in $H$ \\
$\mathcal{H}$ & Hypothesis graph set \\
$Z$ & A reference graph (the ground-truth) or the set of nodes in it \\
$Z \setminus X$ & Set of non-input nodes in $Z$ \\
$\mathcal{Z}$ & Reference graph set \\
$\mathbb{Z}$ & Set of all possible reference graph sets \\
$h$ & Node in a hypothesis graph \\
$z$ & Node in a reference graph \\
$\epsilon$ & Null component identifier for input nodes \\
$c(h)$ & The component that computes node $h$ (with $\epsilon$ for input nodes) \\
$c(z)$ & The component that computes node $z$ (with $\epsilon$ for input nodes) \\
$v(h)$ & The value of node $h$ \\
$v(z)$ & The value of node $z$ \\
$h = h'$ & Equality of two hypothesis nodes: $c(h) = c(h')$ and $v(h) = v(h')$ \\
$z = z'$ & Equality of two reference nodes: $c(z) = c(z')$ and $v(z) = v(z')$ \\
$\mathbf{h}$ & Input nodes to node $h$ \\
$\mathbf{z}$ & Input nodes to node $z$ \\
$h_i$ & $i$-th node in the inputs $\mathbf{h}$ \\
$z_i$ & $i$-th node in the inputs $\mathbf{z}$ \\
$\mathbf{h} = \mathbf{h}'$ & $c(h_i) = c(h_i')$ and $h_i = h_i'$ for all $i$ \\
$\mathbf{z} = \mathbf{z}'$ & $c(z_i) = c(z_i')$ and $z_i = z_i'$ for all $i$ \\
$n$ & Number of input nodes to a single component \\
$\cong_S$ & Structural alignment between $\mathcal{H}$ and $\mathcal{Z}$ \\
$\varphi$ & Bijection between components of $\mathcal{H}$ and $\mathcal{Z}$ for structural alignment \\
$\mathcal{C}$ & Set of all non-output components used in a graph set on training data \\
$c$ & A component in the non‑output component set $\mathcal{C}$ \\
$\xi_c$ & Mapping from hypothesis to reference values for $c \in \mathcal{C}$ on training data \\
\hline
\multicolumn{2}{c}{\textit{Regularization and Mathematical Operations} (Section~\ref{section:example_approach})} \\
$\alpha$ & Variance of Gaussian noise in entropy minimization regularization \\
$\beta$ & $L_2$ norm penalty coefficient in entropy minimization regularization \\
$\mathcal{L}$ & Original training loss function \\
$\mathcal{L}'$ & Total loss with entropy minimization regularization \\
\end{tabularx}
\end{center}

\twocolumn

\section{Proofs}
\label{section:proofs}
\subsection{Proofs in Derivations Section}
\subsubsection{Mathematical Preliminaries}

\begin{replemma}{lemma:mappings_on_nodes}[Mappings on Nodes]
Let \(S_1\) and \(S_2\) be finite sets, with \(S_2\) fixed. A well-defined and surjective mapping \(f: S_1 \to S_2\) is injective if and only if \(|S_1|\) is minimized, i.e., \(|S_1| = |S_2|\).
\end{replemma}
\begin{proof}
\((\Rightarrow)\) Suppose \(f\) is injective. Since \(f\) is well-defined and surjective, it is bijective, so \(|S_1| = |S_2|\). Thus \(|S_1|\) is minimized, as any well-defined and surjective mapping from a smaller domain to \(S_2\) would violate the pigeonhole principle.

\((\Leftarrow)\) We prove the contrapositive. Suppose \(f\) is not injective. Then there exist distinct \(x_1, x_2 \in S_1\) with \(f(x_1) = f(x_2) = y\) for some \(y \in S_2\). Let \(S_1' = S_1 \setminus \{x_2\}\) and let \(f' = f|_{S_1'}\). Since \(f\) is well-defined, \(f'\) is well-defined; since \(y\) still has a preimage \(x_1\) in \(S_1'\), \(f'\) is surjective. But \(|S_1'| = |S_1| - 1 < |S_1|\), so \(|S_1|\) is not minimized. By contraposition, if \(|S_1|\) is minimized, then \(f\) is injective.
Thus the equivalence holds.
\end{proof}
\subsubsection{Necessity}

\begin{repproposition}{proposition:necessity}[Necessity]
A model provably enables compositional generalization (Definition~\ref{definition:compositional_generalization}) only if it has AS-UMR (Definition~\ref{definition:aligned_structure-unambiguous_minimized_representation}).
 
\end{repproposition}
\begin{proof}
Assume the model provably enables compositional generalization. By Assumption~\ref{assumption:correct_training_predictions}, training predictions are correct; by Definition~\ref{definition:compositional_generalization}, test predictions are also correct. Hence, all graphs produce correct outputs. By Assumption~\ref{assumption:seen_test_component_inputs}, all test component inputs have been seen in training. Therefore, the hypothesis graph set $\mathcal{H}$ itself satisfies the conditions of a reference graph set (Definition~\ref{definition:reference_graph_set}). Taking $\mathcal{Z} = \mathcal{H}$ yields structural alignment (Definition~\ref{definition:structure_alignment}), unambiguous representation (Definition~\ref{definition:unambiguous_representation}), and injectivity trivially. Hence, minimized representation (Definition~\ref{definition:minimized_representation}) follows from Lemma~\ref{lemma:mappings_on_nodes}. Thus $\mathcal{H}$ satisfies AS-UMR.
\end{proof}
\subsubsection{Sufficiency}

\begin{lemma}[Surjective Mapping]
\label{lemma:surjective_mapping}
Given that $\mathcal{H} \cong_S \mathcal{Z}$ (Definition~\ref{definition:structure_alignment}), $\forall c \in \mathcal{C}, \xi_c$ is surjective (on the training data).
\end{lemma}
\begin{proof}
The reference graph set (Definition~\ref{definition:reference_graph_set}) is constructed from the training data so that for every possible reference value, there exists at least one training sample that produces it. Hence, the mappings are surjective.
\end{proof}

\begin{lemma}[Injective Component Outputs]
\label{lemma:injective_component_outputs}
Suppose $\mathcal{H}$ is a hypothesis graph set, and there exists a reference graph set $\mathcal{Z} \in \mathbb{Z}$ such that $\mathcal{H} \cong_S \mathcal{Z}$. Then, for every $c \in \mathcal{C}$, if $\mathcal{H}$ has an unambiguous representation (Definition~\ref{definition:unambiguous_representation}) and a minimized representation (Definition~\ref{definition:minimized_representation}) w.r.t.\ $\mathcal{Z}$ at $c$, then $\xi_c$ is injective.
\end{lemma}
\begin{proof}
The graph sets have structural alignment.
For all \(c \in \mathcal{C}\), by Lemma~\ref{lemma:surjective_mapping}, the mapping \(\xi_c\) is surjective.
By the unambiguous representation, it is well-defined.
By the minimized representation and Lemma~\ref{lemma:mappings_on_nodes}, the mapping \(\xi_c\) is injective.
\end{proof}

\begin{lemma}[Component Input]
\label{lemma:component_input}
For a hypothesis graph set $\mathcal{H}$, suppose the following two conditions hold:
\begin{align*}
    (1) \;\; & \exists \mathcal{Z} \in \mathbb{Z}: \mathcal{H} \cong_S \mathcal{Z}, \text{ and}\\
    (2) \;\; & \forall c \in \mathcal{C}, \xi_c \text{ is injective.}
\end{align*}
For all $B \in \mathcal{D}_{\mathrm{test}}, \forall z^B \in Z^B \setminus X^B$,
\begin{align*}
\text{if} \;\; (3) \;\;
    & \forall i \in \{1, \dots, n\}, \\
    & \exists C_i \in \mathcal{D}_{\mathrm{train}}, \exists z^{C_i} \in Z^{C_i}: \\
    & z^{C_i} = z^B_i, h^{C_i} = h^B_i, \\
\text{then} \quad\quad\;\;
    & \exists A \in \mathcal{D}_{\mathrm{train}}, \exists z^A \in Z^A \setminus X^A: \\
    & \mathbf{z}^A = \mathbf{z}^B, \mathbf{h}^A = \mathbf{h}^B.
\end{align*}
\end{lemma}
\begin{proof}
\begin{align*}
\text{For all } B \in \mathcal{D}_{\mathrm{test}}, \forall z^B \in Z^B \setminus X^B,
\end{align*}
there is a training reference component input because $\mathcal{Z}$ is a reference graph set (Definition~\ref{definition:reference_graph_set}) by condition (1):
\begin{align*}
    \exists A \in \mathcal{D}_{\mathrm{train}}, \exists z^A \in Z^A \setminus X^A: \mathbf{z}^A = \mathbf{z}^B.
\end{align*}
It follows that:
\begin{align*}
    \mathbf{z}^A = \mathbf{z}^B \implies \forall i = 1, \dots n: z^A_i = z^B_i.
\end{align*}
For any input node, by condition (3):
\begin{align*}
    \exists C_i \in \mathcal{D}_{\mathrm{train}}, \exists z^{C_i} \in Z^{C_i}:
    z^{C_i} = z^B_i, h^{C_i} = h^B_i.
\end{align*}
Therefore:
\begin{align*}
    z^{C_i} = z^B_i = z^A_i.
\end{align*}
Since $A$ and $C$ are training samples, condition (2) gives the following implication for non‑input nodes; for input nodes, it follows directly from the equality of the inputs (i.e., for input nodes, $z=x=h$):
\begin{align*}
z^{C_i} = z^A_i \implies h^{C_i} = h^A_i.
\end{align*}
Therefore:
\begin{align*}
    h^A_i = h^{C_i} = h^B_i.
\end{align*}
It applies to all input nodes:
\begin{align*}
    \forall i = 1, \dots, n: h^A_i = h^B_i \implies 
    \mathbf{h}^A = \mathbf{h}^B.
\end{align*}
Therefore:
\begin{align*}
    \mathbf{z}^A = \mathbf{z}^B, \mathbf{h}^A = \mathbf{h}^B.
\end{align*}
\end{proof}

\begin{replemma}{lemma:inductive_step}[Inductive Step]
For a hypothesis graph set $\mathcal{H}$, suppose the following two conditions hold:
\begin{align*}
    (1) \;\; & \exists \mathcal{Z} \in \mathbb{Z}: \mathcal{H} \cong_S \mathcal{Z}, \text{ and}\\
    (2) \;\; & \forall c \in \mathcal{C}, \xi_c \text{ is injective.}
\end{align*}
For all $B \in \mathcal{D}_{\mathrm{test}}, \forall z^B \in Z^B \setminus X^B$,
\begin{align*}
\text{if} \;\; (3) \;\;
    & \forall i \in \{1, \dots, n\}, \\
    & \exists C_i \in \mathcal{D}_{\mathrm{train}}, \exists z^{C_i} \in Z^{C_i}: \\
    & z^{C_i} = z^B_i, h^{C_i} = h^B_i, \\
\text{then} \quad\quad\;\;
    & \exists A \in \mathcal{D}_{\mathrm{train}}, \exists z^A \in Z^A \setminus X^A: \\
    & z^A = z^B, h^A = h^B.
\end{align*}

\end{replemma}
\begin{proof}
Lemma~\ref{lemma:component_input} applies:
\begin{align*}
    & \forall B \in \mathcal{D}_{\mathrm{test}}, \forall z^B \in Z^B \setminus X^B, \\
    & \exists A \in \mathcal{D}_{\mathrm{train}}, \exists z^A \in Z^A \setminus X^A: \\
    & \mathbf{z}^A = \mathbf{z}^B, \mathbf{h}^A = \mathbf{h}^B.
\end{align*}
By the deterministic property of components (Definition~\ref{definition:component}):
\begin{align*}
    & \mathbf{z}^A = \mathbf{z}^B \implies z^A = z^B, \\
    & \mathbf{h}^A = \mathbf{h}^B \implies h^A = h^B.
\end{align*}
Therefore:
\begin{align*}
    z^A = z^B, h^A = h^B.
\end{align*}
\end{proof}

\begin{lemma}[Induction over the Graph]
\label{lemma:induction_over_the_graph}
If a model has structural alignment (Definition~\ref{definition:structure_alignment}) and $\xi_c$ is injective for all $c \in \mathcal{C}$, then:
\begin{align*}
& \forall B \in \mathcal{D}_{\mathrm{test}}, \forall z^B \in Z^B, \exists A \in \mathcal{D}_{\mathrm{train}}, \exists z^A \in Z^A: \\
&    z^A = z^B, h^A = h^B.
\end{align*}
\end{lemma}
\begin{proof}
We proceed by induction over the nodes in topological order.
$\forall B \in \mathcal{D}_{\mathrm{test}}$:
\subsubsection*{Base step}
For any test input node \(x^B \in X^B\), by Definition~\ref{definition:computational_graph}, it must serve as an input to at least one non-input component (otherwise it would be a dangling node). Let \(z^B\) be such a non-input component output node. Since \(\mathcal{Z}\) is a reference graph set, Definition~\ref{definition:reference_graph_set} applies to \(z^B\): there exists a training sample \(A\) and a corresponding node \(z^A\) such that \(\mathbf{z}^A = \mathbf{z}^B\). Since \(x^B\) is in the tuple \(\mathbf{z}^B\), the equality of tuples implies that the corresponding input node in \(\mathbf{z}^A\), denoted \(x^A\), satisfies \(x^A = x^B\). Hence, the base step holds for all graph input nodes:
\begin{align*}
    & \forall z^B \in X^B, \exists A \in \mathcal{D}_{\mathrm{train}}, \exists z^A \in X^A: \\
    & z^A = x^A = x^B = z^B, \quad
     h^A = x^A = x^B = h^B
\end{align*}
\subsubsection*{Inductive step}
Assume the property holds for all input nodes of the current node. Then condition (3) of Lemma~\ref{lemma:inductive_step} is satisfied. Since $\xi_c$ is injective for all $c \in \mathcal{C}$, condition (2) of Lemma~\ref{lemma:inductive_step} holds as well. Structural alignment gives condition (1). Therefore, by Lemma~\ref{lemma:inductive_step}, the property holds for the output node of the component.

By induction, the property holds for every node:
\begin{align*}
& \forall z^B \in Z^B, \exists A \in \mathcal{D}_{\mathrm{train}}, \exists z^A \in Z^A: \\
&    z^A = z^B, h^A = h^B.
\end{align*}
\end{proof}

\begin{lemma}[Injective Mapping for Sufficiency]
\label{lemma:injective_mapping_for_sufficiency}
A model provably enables compositional generalization (Definition~\ref{definition:compositional_generalization}) if it has structural alignment (Definition~\ref{definition:structure_alignment}) and $\xi_c$ is injective for all $c \in \mathcal{C}$.
\end{lemma}
\begin{proof}
By Lemma~\ref{lemma:induction_over_the_graph}:
\begin{align*}
& \forall B \in \mathcal{D}_{\mathrm{test}}, \forall z^B \in Z^B, \exists A \in \mathcal{D}_{\mathrm{train}}, \exists z^A \in Z^A: \\
&    z^A = z^B, h^A = h^B.
\end{align*}
It includes output nodes $Y^B \subseteq Z^B$, therefore:
\begin{align*}
    y^A = y^B, \hat{y}^A = \hat{y}^B.
\end{align*}
$y^A$ is also an output node, because $y^A$ and $y^B$ correspond to the same component, and whether they are output nodes is consistent.
By the premise of compositional generalization (Definition~\ref{definition:compositional_generalization}), training predictions are correct:
\begin{align*}
    \forall y^A \in Y^A: \hat{y}^A = y^A.
\end{align*}
Therefore:
\begin{align*}
    \hat{y}^B = \hat{y}^A = y^A = y^B.
\end{align*}
It applies to all output nodes.
Therefore:
\begin{align*}
    \hat{Y}^B = Y^B.
\end{align*}
By Definition~\ref{definition:compositional_generalization}, compositional generalization is enabled.
\end{proof}

\begin{repproposition}{proposition:sufficiency}[Sufficiency]
A model provably enables compositional generalization (Definition~\ref{definition:compositional_generalization}) if it has AS-UMR (Definition~\ref{definition:aligned_structure-unambiguous_minimized_representation}).

\end{repproposition}
\begin{proof}
Lemma~\ref{lemma:injective_component_outputs} shows that under the conditions of AS-UMR (Definition~\ref{definition:aligned_structure-unambiguous_minimized_representation}), the mapping has an injective representation for all $c \in \mathcal{C}$. By Lemma~\ref{lemma:injective_mapping_for_sufficiency}, compositional generalization holds.
\end{proof}
\subsubsection{Theorem}

\begin{reptheorem}{theorem:necessary_and_sufficient_condition}[Necessary and Sufficient Condition]
A model provably enables compositional generalization (Definition~\ref{definition:compositional_generalization}) if and only if it has Aligned Structure-Unambiguous Minimized Representation (AS-UMR, Definition~\ref{definition:aligned_structure-unambiguous_minimized_representation}).

\end{reptheorem}
\begin{proof}
    Necessity is given by Proposition~\ref{proposition:necessity}; sufficiency by Proposition~\ref{proposition:sufficiency}.
\end{proof}

\subsection{Proofs in Example Approach Section}
\begin{lemma}[Entropy Decreases Under Event Merging]
\label{lemma:entropy_decreases_under_event_merging}
Let \(P = (p_1, p_2, \ldots, p_m)\) be a probability distribution on \(m \ge 2\) elements, where \(p_k > 0\) for all \(k\). Pick any two distinct indices \(i, j \in \{1, \ldots, m\}\). Define the probability distribution \(Q\) on \(m - 1\) elements by merging these two specific events:
\begin{align*}
Q = (p_1, \ldots, p_{i-1}, p_{i+1}, \ldots, p_{j-1}, \\
p_{j+1}, \ldots, p_m, \; p_i + p_j),    
\end{align*}
i.e., the probabilities of the two chosen events are summed into a single event, while all other events remain unchanged. Then the entropy strictly decreases:
\[
H(Q) < H(P).
\]
\end{lemma}
\begin{proof}
Since entropy is invariant under permutations of event labels, without loss of generality, we relabel the indices so that the two merged events are the first two, with probabilities \(p_1\) and \(p_2\). Under this relabeling, we have:
\begin{align*}
& P = (p_1, p_2, p_3, \ldots, p_m), \\
& Q = (p_1 + p_2, p_3, p_4, \ldots, p_m).    
\end{align*}
By the definition of entropy:
\begin{align*}
H(P) & = -\sum_{i=1}^n p_i \log_2 p_i, \\
H(Q) & = -(p_1 + p_2)\log_2(p_1 + p_2) - \sum_{i=3}^n p_i \log_2 p_i.
\end{align*}
Subtract \( H(Q) \) from \( H(P) \); the terms for \( i \geq 3 \) cancel out exactly:
\begin{align*}
& H(P) - H(Q) \\
= & -p_1 \log_2 p_1 - p_2 \log_2 p_2 + (p_1 + p_2)\log_2(p_1 + p_2).
\end{align*}
Rearrange terms to isolate the binary distribution contribution:
\begin{align*}
& H(P) - H(Q) \\
= & p_1 \log_2\left(\frac{p_1 + p_2}{p_1}\right) + p_2 \log_2\left(\frac{p_1 + p_2}{p_2}\right).
\end{align*}
Factor out \( p_1 + p_2 \) to reveal the binary entropy:
\[
H(P) - H(Q) = (p_1 + p_2) \cdot H\left( \frac{p_1}{p_1 + p_2}, \frac{p_2}{p_1 + p_2} \right),
\]
where \( H(a, 1-a) = -a\log_2 a - (1-a)\log_2(1-a) \) is the entropy of a binary distribution.

Since \( p_1 > 0 \) and \( p_2 > 0 \), both probabilities in the binary distribution are strictly between 0 and 1, so \( H\left( \frac{p_1}{p_1 + p_2}, \frac{p_2}{p_1 + p_2} \right) > 0 \). Also, \( p_1 + p_2 > 0 \). Therefore:
\[
H(P) - H(Q) > 0 \implies H(Q) < H(P),
\]
which proves that merging any two distinct events with positive probability strictly reduces the entropy. 
\end{proof}

\begin{replemma}{lemma:minimum_entropy_implies_minimized_representation}[Minimum Entropy Implies Minimized Representation]
Given that $\mathcal{H} \cong_S \mathcal{Z}$ (Definition~\ref{definition:structure_alignment}) and $\mathcal{H}$ has an unambiguous representation (Definition~\ref{definition:unambiguous_representation}) w.r.t.\ $\mathcal{Z}$ at $c \in \mathcal{C}$, if $c$ has minimum entropy (Definition~\ref{definition:component-wise_minimum_entropy}), then $\mathcal{H}$ has a minimized representation (Definition~\ref{definition:minimized_representation}) w.r.t.\ $\mathcal{Z}$ at $c$.
\end{replemma}
\begin{proof}
Fix an arbitrary \(c \in \mathcal{C}\). Assume for contradiction that the representation is not minimized. Let \(D \triangleq \operatorname{domain}(\xi_c)\) and \(R \triangleq \operatorname{image}(\xi_c)\). Since the representation is unambiguous, the mapping \(\xi_c\) is well-defined. By Lemma~\ref{lemma:surjective_mapping}, it is surjective. Since the representation is not minimized, \(|D| > |R|\). By the pigeonhole principle, there exist two distinct hypothesis values \(h_1 \neq h_2 \in D\) that map to the same reference value \(z \in R\). Merging these two events in the probability distribution over \(D\) strictly reduces its entropy (Lemma~\ref{lemma:entropy_decreases_under_event_merging}), contradicting the assumption that \(c\) has minimum entropy. Therefore, the representation must be minimized.
\end{proof}

\begin{repcorollary}{corollary:example_approach_verification}[Example Approach Verification]
If component-wise minimum entropy (Definition~\ref{definition:component-wise_minimum_entropy}) is achieved, a model trained according to Algorithm~\ref{algorithm:example_approach} enables compositional generalization (Definition~\ref{definition:compositional_generalization}).
\end{repcorollary}
\begin{proof}
    By the architectural design, structure alignment (Definition~\ref{definition:structure_alignment}) holds.
    By the training data preparation, the unambiguous representation (Definition~\ref{definition:unambiguous_representation}) holds for each component.
    Since component-wise minimum entropy holds, by Lemma~\ref{lemma:minimum_entropy_implies_minimized_representation}, the minimized representation (Definition~\ref{definition:minimized_representation}) holds for each component. 
    Thus, AS-UMR (Definition~\ref{definition:aligned_structure-unambiguous_minimized_representation}) is satisfied; hence, by Proposition~\ref{proposition:sufficiency}, compositional generalization holds.
\end{proof}

\subsection{Proofs in Minimal Example Section}
\begin{replemma}{lemma:unambiguous_representation_verification}[Unambiguous Representation Verification]
Given correct training predictions, the training samples shown in Table~\ref{tab:values_for_the_example} have an unambiguous representation (Definition~\ref{definition:unambiguous_representation}) for every non-output component.
\end{replemma}
\begin{proof}
According to Algorithm~\ref{algorithm:forward_pass_for_the_model}, all samples share the same graph structure. We only need to verify the intermediate node $z$, as it is the output of the only non-output component. Hence, to prove:
\begin{align*}
    \forall A, C \in \mathcal{D}_{\mathrm{train}}: h^A = h^C \implies z^A = z^C.
\end{align*}

If $A$ and $C$ are the same sample, then $z^A = z^C$, so the conclusion holds. Assume they are different samples.

We will first check which pairs can have the same hypothesis, and then whether they have the same reference.
We use the property of deterministic neural networks (Definition~\ref{definition:component}):
\begin{align*}
    x^A_1 = x^C_1, x^A_2 = x^C_2 \implies h^A = h^C.
\end{align*}
From Table~\ref{tab:values_for_the_example}:
\begin{align*}
    h^a = h^e, h^b = h^f.
\end{align*}
By deterministic neural network (Definition~\ref{definition:component}) and correct training predictions:
\begin{align*}
    h^A = h^C, x^A_3 = x^C_3 \implies \hat{y}^A = \hat{y}^C \implies y^A = y^C.
\end{align*}
Taking the contrapositive gives:
\begin{align*}
    x^A_3 = x^C_3, y^A \neq y^C \implies h^A \neq h^C.
\end{align*}
Hence, from Table~\ref{tab:values_for_the_example}:
\begin{align*}
    h^a \neq h^b, h^c \neq h^d, h^c \neq h^e, h^d \neq h^f, h^e \neq h^f.
\end{align*}
We use the equalities to extend the non-equalities:
\begin{align*}
    h^a=h^e,h^c \neq h^e & \implies h^a \neq h^c, \\
    h^a=h^e,h^e \neq h^f & \implies h^a \neq h^f, \\
    h^b=h^f,h^d \neq h^f & \implies h^b \neq h^d, \\
    h^b=h^f,h^e \neq h^f & \implies h^b \neq h^e.
\end{align*}
This yields the remaining six possible equal pairs:
\begin{align*}
    (a, d), (a, e), (b, c), (b, f), (c, f), (d, e).
\end{align*}
By the definition of $z$ in Algorithm~\ref{algorithm:ground-truth_function}, $z = x_1 \oplus x_2$.
Hence, as shown in Table~\ref{tab:values_for_the_example}, in all these pairs, the two samples have the same $z$ value:
\begin{align*}
    z^a=z^d, z^a=z^e, z^b=z^c, z^b=z^f, z^c=z^f, z^d=z^e.
\end{align*}
Therefore:
\begin{align*}
    h^A = h^C \implies z^A = z^C.
\end{align*}
\end{proof}

\begin{repcorollary}{corollary:minimal_example_verification}[Minimal Example Verification]
In the minimal example task (Section~\ref{section:task}), if component-wise minimum entropy (Definition~\ref{definition:component-wise_minimum_entropy}) is achieved, the algorithm designed in Section~\ref{section:xor_algorithm} enables compositional generalization (Definition~\ref{definition:compositional_generalization}).
\end{repcorollary}
\begin{proof}
The design in Section~\ref{section:xor_algorithm} follows Algorithm~\ref{algorithm:example_approach}. In particular, Lemma~\ref{lemma:unambiguous_representation_verification} establishes the unambiguous representation condition. By Corollary~\ref{corollary:example_approach_verification}, the conclusion holds.

Note that, in this problem, Lemma~\ref{lemma:unambiguous_representation_verification} establishes unambiguous representation under correct training predictions alone, without requiring the component-wise minimum entropy condition in Algorithm~\ref{algorithm:example_approach}.
\end{proof}

\subsection{Proofs in Discussion Section}
\begin{lemma}[Seen Test Component Inputs]
\label{lemma:seen_test_component_inputs}
For a hypothesis graph set $\mathcal{H}$, if there exists a reference graph set $\mathcal{Z} \in \mathbb{Z}$:
\begin{enumerate}
    \item $\mathcal{H} \cong_S \mathcal{Z}$ (Definition~\ref{definition:structure_alignment}); and
    \item for every $c \in \mathcal{C}$, $\mathcal{H}$ has, w.r.t.\ $\mathcal{Z}$ at $c$:
    \begin{enumerate}
        \item an unambiguous representation (Definition~\ref{definition:unambiguous_representation}), and
        \item a minimized representation (Definition~\ref{definition:minimized_representation}).
    \end{enumerate}
\end{enumerate}
then all test component inputs are seen in training:
\begin{align*}
    & \forall B \in \mathcal{D}_{\mathrm{test}}, \forall h^B \in H^B \setminus X^B, \\
    & \exists A \in \mathcal{D}_{\mathrm{train}}, \exists h^A \in H^A \setminus X^A: \mathbf{h}^A = \mathbf{h}^B.
\end{align*}
\end{lemma}
\begin{proof}
Because of structural alignment (Definition~\ref{definition:structure_alignment}), unambiguous representation (Definition~\ref{definition:unambiguous_representation}), and minimized representation (Definition~\ref{definition:minimized_representation}),
Lemma~\ref{lemma:injective_component_outputs} holds for all $c \in \mathcal{C}$, and then Lemma~\ref{lemma:induction_over_the_graph} holds. 
Lemma~\ref{lemma:component_input} holds because its three conditions hold: (1) by Definition~\ref{definition:structure_alignment}, (2) by Lemma~\ref{lemma:injective_component_outputs} for all $c \in \mathcal{C}$, (3) by Lemma~\ref{lemma:induction_over_the_graph}.
Thus:
\begin{align*}
    & \forall B \in \mathcal{D}_{\mathrm{test}}, \forall h^B \in H^B \setminus X^B, \\
    & \exists A \in \mathcal{D}_{\mathrm{train}}, \exists h^A \in H^A \setminus X^A: \\
    & \mathbf{z}^A = \mathbf{z}^B, \mathbf{h}^A = \mathbf{h}^B 
    \implies \mathbf{h}^A = \mathbf{h}^B.
\end{align*}
\end{proof}

\begin{reptheorem}{theorem:alternative_necessary_and_sufficient_condition}[Alternative Necessary and Sufficient Condition]
A model enables alternative compositional generalization (Definition~\ref{definition:alternative_compositional_generalization}) if and only if it has AS-UMR (Definition~\ref{definition:aligned_structure-unambiguous_minimized_representation}).

\end{reptheorem}
\begin{proof}
We prove necessity and sufficiency.

\subsubsection*{Necessity}
By Definition~\ref{definition:alternative_compositional_generalization}, the alternative compositional generalization implies both compositional generalization (Definition~\ref{definition:compositional_generalization}) and seen test component inputs (corresponding to Assumption~\ref{assumption:seen_test_component_inputs}).
Therefore, by Proposition~\ref{proposition:necessity}, the conclusion holds.

\subsubsection*{Sufficiency}
By Proposition~\ref{proposition:sufficiency}, the model enables compositional generalization (Definition~\ref{definition:compositional_generalization}).
Moreover, by Lemma~\ref{lemma:seen_test_component_inputs}, the test component inputs are seen in training.
Hence, by Definition~\ref{definition:alternative_compositional_generalization}, the model enables alternative compositional generalization.
\end{proof}

\section{Experiment Details}
\label{section:experiment_details}
\subsection{Details of Settings}
We use TensorFlow~\cite{tensorflow2015-whitepaper} for implementation.

The embedding size is 64.
Each feed-forward neural network has two hidden layers, each with 64 nodes and ReLU activation.
When using entropy minimization regularization (Algorithm~\ref{algorithm:entropy_minimization_regularization}), it is uniformly applied to all the layers in a feed-forward network.
We use cross-entropy for prediction loss.
$\alpha$ is 0.01 and $\beta$ is 0.01.
We use the Adam optimizer with default settings and a learning rate of 0.001.
The batch size is 1,000.
The models are trained for 1,000 iterations.

For the baseline model, the embeddings of $x_1, x_2, x_3$ are concatenated and flattened into a single vector before being fed into a feed-forward neural network with two hidden layers, each with 256 nodes and ReLU activation. Other hyperparameters do not change.
  
\subsection{Robustness to Hyperparameter Variations}
We tune parameters for $(\alpha, \beta)$ over the grid $\{0.1, 0.01, 0.001\}\times\{0.1, 0.01, 0.001\}$.
For all nine combinations, the resulting accuracy is 1.0 $\pm$ 0.0 on both the training and test datasets, supporting robustness to hyperparameter variations.

\section{Additional Discussion}
\label{section:additional_discussion}
\subsection{Intuition Behind the Inductive Step}
\label{sec:intuition_behind_the_inductive_step}
We develop intuition for the key issue that the inductive step (Lemma~\ref{lemma:inductive_step}) addresses. We start from a familiar single-component in-domain setting and then extend the reasoning to multi-component compositional generalization.

\subsubsection*{Single-Component Network}
Consider standard in-domain generalization with a single-component network. Let $(X^A, Y^A)$ be a training sample and $(X^B, Y^B)$ a test sample with $X^A = X^B$. The correctness of the test prediction follows from: (1) the ground-truth mapping is deterministic, so $Y^A = Y^B$; and (2) the network is deterministic, so $\hat{Y}^A = \hat{Y}^B$. Since training predictions are correct ($\hat{Y}^A = Y^A$), we conclude:
\[
\hat{Y}^B = \hat{Y}^A = Y^A = Y^B.
\]

\subsubsection*{Rewriting the Single-Component Case in Our Notation}
A single-component network is a special case of a multi-component network. In our notation:
\[
\mathbf{z} = X, \quad \mathbf{h} = X, \quad z = Y, \quad h = \hat{Y}.
\]
Crucially, the hypothesis input $\mathbf{h}$ and the reference input $\mathbf{z}$ coincide because both equal the model input. Thus:
\[
\mathbf{h}^A = X^A = \mathbf{z}^A = \mathbf{z}^B = X^B = \mathbf{h}^B.
\]
Paralleling the single-component case, we have:
\[
z^A = z^B \text{ and } h^A = h^B.
\]

\subsubsection*{Multi-Component Network}
For a multi-component network, a component input may no longer be the model input. Consider a test sample $B$ and a node $z^B \in Z^B$. By Definition~\ref{definition:reference_graph_set}, we can select a training sample $A$ and a node $z^A \in Z^A$ such that $\mathbf{z}^A = \mathbf{z}^B$. However, $\mathbf{h}^A = \mathbf{h}^B$ may not hold. We therefore require:
\[
\mathbf{z}^A = \mathbf{z}^B \implies \mathbf{h}^A = \mathbf{h}^B.
\]
This is the key to the inductive step, and the conditions in Lemma~\ref{lemma:inductive_step} guarantee it.
\subsection{Injective Representation}
\label{sec:injective_representation}
We discuss an equivalent formulation of AS-UMR (Definition~\ref{definition:aligned_structure-unambiguous_minimized_representation}).
\begin{definition}[Injective Representation]
\label{definition:injective_representation}
Given $\mathcal{H} \cong_S \mathcal{Z}$ (Definition~\ref{definition:structure_alignment}), $\mathcal{H}$ has an \textit{injective representation} w.r.t.\ $\mathcal{Z}$ at $c \in \mathcal{C}$ if the component-wise mapping $\xi_c$ is injective---i.e., no two distinct hypothesis values map to the same reference value on training data.
\end{definition}

\begin{definition}[Aligned Structure--Injective Representation, AS-IR]
\label{definition:aligned_structure-injective_representation}
A model with hypothesis graph set $\mathcal{H}$ satisfies \textit{AS-IR} if there exists a reference graph set $\mathcal{Z} \in \mathbb{Z}$ such that:
\begin{enumerate}
    \item $\mathcal{H}$ and $\mathcal{Z}$ have structural alignment (Definition~\ref{definition:structure_alignment}), i.e., $\mathcal{H} \cong_S \mathcal{Z}$;
    \item for every $c \in \mathcal{C}$, $\mathcal{H}$ has an injective representation w.r.t.\ $\mathcal{Z}$ at $c$ (Definition~\ref{definition:injective_representation}).
\end{enumerate}
\end{definition}

This yields an equivalent characterization:
\begin{corollary}[Necessary and Sufficient Condition via Injectivity]
A model provably enables compositional generalization (Definition~\ref{definition:compositional_generalization}) if and only if it has AS-IR (Definition~\ref{definition:aligned_structure-injective_representation}).
\end{corollary}

\begin{proof}
\textbf{Necessity.} By Proposition~\ref{proposition:necessity}, compositional generalization implies AS-UMR. By Lemma~\ref{lemma:injective_component_outputs} for all $c \in \mathcal{C}$, AS-UMR implies AS-IR.

\textbf{Sufficiency.} By Lemma~\ref{lemma:injective_mapping_for_sufficiency}, the conclusion holds.
\end{proof}

Intuitively, given structural alignment, an injective representation is equivalent to the conjunction of an unambiguous representation and a minimized representation. The key observation is that the reference graph set $\mathcal{Z}$ needs only to exist (Definition~\ref{definition:aligned_structure-unambiguous_minimized_representation}), so we are free to choose a $\mathcal{Z}$ such that the number of distinct reference outputs per non-output component is minimized.

Nevertheless, Theorem~\ref{theorem:necessary_and_sufficient_condition} has two advantages. First, the decomposition into an unambiguous and minimized representation is more interpretable: it captures the idea of ``sufficient but not excessive information''. Second, it directly suggests algorithmic design levers---diverse training data and entropy minimization regularization (Section~\ref{sec:approach}).

\section{Comparison with Related Work}
\label{section:comparison_with_related_work}
Among recent related theoretical work, we compare those that are more closely related to ours.

\subsection{Jarvis et al. (2023)}
Our necessary and sufficient condition (Theorem~\ref{theorem:necessary_and_sufficient_condition}) complements the findings of \citet{jarvis2023on}, who demonstrated that only fully partitioned architectures, which strictly segregate compositional and non-compositional features, can achieve systematic generalization. Their result can be viewed as a special instantiation of our structural alignment condition (Definition~\ref{definition:structure_alignment}), where the perfect partitioning ensures that the computational graph matches the true compositional hierarchy. However, our framework extends beyond modular architectures by providing the unambiguous and minimized representation conditions, which characterize the required properties of training data and regularization.

\subsection{Wiedemer et al. (2023)}
\citet{NEURIPS2023_15f6a108} provide sufficient conditions for compositional generalization in a supervised regression setting under the assumption that the composition function $C$ is known and differentiable, and that the training distribution has compositional support and sufficient support (a rank condition on the Jacobian of $C$). 
We note that when a compositional data-generating process exists and their sufficient support condition is satisfied, our AS-UMR condition is typically implied by their assumptions.
On the other hand, our Theorem~\ref{theorem:necessary_and_sufficient_condition} gives a necessary and sufficient condition that applies to arbitrary neural networks without assuming a known $C$ or differentiability.

\subsection{Ahuja and Mansouri (2024)}
\citet{ahuja2024on} establish provable guarantees for length and compositional generalization in sequence-to-sequence models (deep sets, transformers, SSMs, RNNs) under structural assumptions such as realizability, regular closed supports, and token diversity. Their result implies that, when these conditions hold, the learned representations linearly identify the true latent representations, which in turn ensures generalization to unseen lengths and token combinations. In the context of our necessary and sufficient condition (Theorem~\ref{theorem:necessary_and_sufficient_condition}), their assumptions align with our unified inductive bias: structural alignment is encoded by the chosen architecture, while unambiguous and minimized representations are achieved through data diversity and capacity constraints.
Conversely, our condition operates under a distinct set of assumptions, as it does not require invertible activations, regular closed supports, or a known composition function.

\subsection{Lippl and Stachenfeld (2025)}
\citet{lippl2025when} provide an analysis of compositional generalization in kernel models, identifying conjunction-wise additivity as a key constraint and revealing failure modes such as memorization leak and shortcut bias. In contrast, this work establishes a necessary and sufficient condition for provable compositional generalization that applies to general neural networks without assuming kernel regimes or compositional input structure. Overall, their findings offer a detailed characterization within a specific kernel regime, while our framework provides complementary guarantees for the broader class of neural networks.

\section{SCAN Jump Task}
\label{sec:scan_jump_task}
We demonstrate why the algorithm of \citet{li2019compositional} enables compositional generalization on the SCAN jump task. To our knowledge, providing such a principled explanation for successful algorithms on this benchmark has not been a central focus of prior theoretical work.

\subsection{Task}
The SCAN dataset~\cite{lake2018generalization} is a standard benchmark for evaluating compositional generalization. It consists of command–action pairs, where each command is a natural-language instruction, and each action is a corresponding sequence of atomic operations. For example, \texttt{run twice} maps to \texttt{RUN RUN}.

\paragraph{Data generation.}
In the jump task (a primitive substitution task), the entire dataset $\mathcal{D}$ can be viewed as generated from templates (sentence schemas with placeholders) and two finite vocabularies:
\begin{center}
    \centering
    \begin{tabular}{ll}
        \textbf{action words}: & \texttt{jump}, \texttt{run}, \texttt{walk}, \texttt{look} \\
        \textbf{direction words}: & \texttt{left}, \texttt{right}
    \end{tabular}
\end{center}
Each template contains one or more slots, each designated for either an action word or a direction word. For example:
\begin{itemize}
\item \texttt{[action]}
\item \texttt{[action] [direction] and [action]}
\end{itemize}
A concrete sample is produced by instantiating every slot with a word from the corresponding vocabulary. The dataset $\mathcal{D}$ includes all possible combinations of templates and vocabulary assignments.

\paragraph{Training and test split.}
The training set $\mathcal{D}_\mathrm{train}$ contains:
\begin{itemize}
    \item all samples that \textbf{do not} contain the word \texttt{jump}, and
    \item the single‑word sample \texttt{jump} (i.e., the primitive in isolation).
\end{itemize}
The test set $\mathcal{D}_\mathrm{test}$ comprises all remaining samples, i.e., those that contain \texttt{jump} in any context other than isolation. Thus, during training, the model sees the primitive \texttt{jump} only as a standalone input, while at test time it must process \texttt{jump} combined with other words (see Table~\ref{table:scan_inputs} for examples). This split enforces the challenge of compositional generalization by combining \texttt{jump} with other words.

\begin{table}[!ht]
\centering
\caption{
Examples of input commands from the SCAN jump task. Training samples (upper block) contain the primitive command \texttt{jump} only in isolation, while test samples (lower block) require compositional generalization by combining \texttt{jump} with other words.
}
\begin{tabular}{l} 
    \texttt{\textbf{jump}} \\
    \texttt{run after run left} \\
    \texttt{look twice and look opposite right} \\
    \hline
    \texttt{\textbf{jump} after run left} \\
    \texttt{run after \textbf{jump} left} \\
    \texttt{\textbf{jump} right after \textbf{jump} left twice} 
\end{tabular}
\label{table:scan_inputs}
\end{table}

\subsection{Algorithm}
\label{section:scan_algorithm}
We now describe the algorithm from \citet{li2019compositional} for the SCAN jump task. We treat the sequence-to-sequence module as consisting of multiple components, each responsible for one output position.

The algorithm appends \texttt{<eos>} to the input and \texttt{EOS} to the output for all samples. The input $X$ is a sequence of $n$ words, and the ground-truth output $Y$ is a sequence of $m$ actions:
\begin{align*}
    X = X_1, \dots, X_n, \qquad Y = Y_1, \dots, Y_m.
\end{align*}
In training, $m$ is known; at test time, the model stops generating when it emits \texttt{EOS}, thereby determining the length of the predicted output.
A syntax embedding \(T_i\) and a semantic embedding \(V_i\) are extracted for each word:
\begin{align*}
    X_i \rightarrow T_i, V_i, \quad \forall i = 1, \dots, n.
\end{align*}
The embedding components are shared across input positions, and the entropy minimization regularization (Algorithm~\ref{algorithm:entropy_minimization_regularization}) is applied to both embedding components. The word embeddings are concatenated to form the sentence syntax embedding \(T\) and the sentence semantic embedding \(V\), respectively:
\[
T_1, \dots, T_n \rightarrow T, \qquad V_1, \dots, V_n \rightarrow V.
\]
From \(T\), a sequence of \(m\) softmax attention maps is generated.
\begin{align*}
    T \rightarrow U_1, \dots, U_m.
\end{align*}
For each output position, the corresponding attention map and the sentence semantic embeddings produce a context vector:
\begin{align*}
    U_j, V \rightarrow C_j, \quad \forall j = 1,\dots, m.
\end{align*}
A shared output component then maps the context vector to an action prediction:
\begin{align*}
    C_j \rightarrow \hat{Y}_j, \quad \forall j = 1,\dots, m.
\end{align*}
The final prediction is the concatenation of the individual action predictions:
\begin{align*}
    \hat{Y}_1, \dots, \hat{Y}_m \rightarrow \hat{Y}.
\end{align*}
This procedure is illustrated in Figure~\ref{fig:scan_algorithm}.
\begin{figure}[!ht]
\centering
\begin{tikzpicture}
  \matrix (m) [matrix of math nodes, row sep=0em, column sep=0em]
  {
     X       & \rightarrow & X_1        & \dots      & X_i        & \dots      & X_n        & \text{input} \\
             &             & \downarrow &            & \downarrow &            & \downarrow & \\
     T       & \leftarrow  & T_1        & \dots      & T_i        & \dots      & T_n        & \text{syntax} \\
             &             & V_1        & \dots      & V_i        & \dots      & V_n        & \text{semantics} \\
             &             & \downarrow &            &            &            &            & \\
     \hat{Y} & \leftarrow  & \hat{Y}_1  & \hat{Y}_2  & \dots      & \hat{Y}_m  &            & \text{output} \\};
  \draw[-latex] (m-3-1) [rounded corners]|- (m-5-3) ;
  \path[-stealth]
    (m-4-3) edge (m-6-6)
    (m-4-7) edge (m-6-4)
    ;
\end{tikzpicture}
\caption{
    Illustration of the SCAN jump task algorithm.
    We apply entropy minimization regularization (Algorithm~\ref{algorithm:entropy_minimization_regularization}) to the syntax $T_i$ and semantics $V_i$ word embeddings.
}
\label{fig:scan_algorithm}
\end{figure}

The algorithm involves the following types of components:
\begin{itemize}
    \item A shared word syntax embedding: $X_i \rightarrow T_i$,
    \item A shared word semantic embedding: $X_i \rightarrow V_i$,
    \item $m$ attention maps: $T \rightarrow U_j$, and
    \item A shared output component: $C_j \rightarrow \hat{Y}_j$.
\end{itemize}

The computational structure itself can be viewed as a program (Remark~\ref{remark:structure_as_a_program}). This program takes the sentence syntax embedding as a control signal and dynamically selects a sequence of attention maps to process the data carried by the semantic embeddings.

\subsection{Analysis}
We discuss why the algorithm enables compositional generalization.
Since softmax tends to focus on one position, we assume that each attention is on a single position. Output length should be determined by syntax, but the algorithm ties it to the semantic embedding (via \texttt{EOS} emission); to simplify the analysis, we assume that training samples with the same sentence syntax embedding produce the same output length (equivalently, we can modify the algorithm to enforce it). The regularization minimizes the semantic embeddings, making them unique for each output primitive. The output component then acts as a bijection between these embeddings and the output tokens; hence, we omit the output component from the following analysis.

We have the following lemma (proof in Section~\ref{sec:equal_action_word_syntax_embedding}).
\begin{lemma}[SCAN Task Algorithm Property]
\label{lemma:scan_task_algorithm_property}
With the algorithm in Section~\ref{section:scan_algorithm}, when correct training predictions hold and the entropies of syntax and semantic representations are minimized, all action words have equal hypothesis syntax embeddings.
\end{lemma}
This means that attention maps do not change when switching action words (e.g., \texttt{look twice} $\rightarrow$ \texttt{run twice}). It also implies that the attention maps attend to the corresponding input positions, and consequently, the test predictions are correct (Lemma~\ref{lemma:scan_generalization}).

\begin{lemma}[SCAN Generalization]
\label{lemma:scan_generalization}
With the algorithm in Section~\ref{section:scan_algorithm}, when the entropies of syntax and semantic representations are minimized, compositional generalization (Definition~\ref{definition:compositional_generalization}) holds.
\end{lemma}
\begin{proof}
Suppose correct training predictions hold. By Lemma~\ref{lemma:scan_task_algorithm_property}, all action words share the same syntax embedding:
\[
h_T^{\text{\texttt{jump}}} = h_T^{\text{\texttt{run}}} = h_T^{\text{\texttt{walk}}} = h_T^{\text{\texttt{look}}}.
\]

Consider two training samples that share the same template but differ only in one action-word position---for instance, \texttt{look twice} and \texttt{run twice}. Because the training predictions are correct and the output must change when that action word changes, each attention map must attend to the corresponding action-word position; otherwise, swapping that word would leave the output unchanged, causing a training error. By the construction of the SCAN jump task, such sample pairs exist for every action-word position. Moreover, the semantic embeddings at the attended positions yield the correct output action on training samples.

Now take any test sample \(B\). For each output position \(j\), let \(A\) be a training sample matching \(B\) except that the action word at the corresponding position is replaced with another action word. The shared syntax embedding (Lemma~\ref{lemma:scan_task_algorithm_property}) implies that \(A\) and \(B\) produce identical attention maps, and since attention maps are correct on training samples, the attention for position \(j\) in \(B\) is also on the correct input position. Moreover, the semantic embedding is correct for words attended in training; hence, it yields the correct output action:
\[
\hat{Y}_j^B = Y_j^B.
\]
Since this holds for every output position \(j=1,\dots,m\), we have \(\hat{Y}^B = Y^B\). As \(B\) was arbitrary, compositional generalization holds.
\end{proof}

\begin{lemma}[SCAN AS-UMR Verification]
If (1) the entropies of syntax and semantic representations are minimized, and (2) correct training predictions hold, then a model trained with the algorithm in Section~\ref{section:scan_algorithm} satisfies AS-UMR (Definition~\ref{definition:aligned_structure-unambiguous_minimized_representation}).
\end{lemma}
\begin{proof}
By (1), Lemma~\ref{lemma:scan_generalization} holds, so compositional generalization holds.
Also, all test component inputs are seen in training:
\begin{enumerate}
    \item The syntax embedding component $X_i \rightarrow T_i$ has seen input because all test input words are seen in training.
    \item Each attention map generation component $T \rightarrow U_j$ has seen input because all action word syntax embeddings are equal (Lemma~\ref{lemma:scan_task_algorithm_property}), and all test combinations of templates and direction words are seen in training.
    \item The semantic embedding component $X_i \rightarrow \hat{Y}_j$ only sees attended words. Since all attentions are correct, all attended test words are seen in training.
\end{enumerate}
Hence, alternative compositional generalization (Definition~\ref{definition:alternative_compositional_generalization}) holds. By (2), the necessity direction of Theorem~\ref{theorem:alternative_necessary_and_sufficient_condition} implies AS-UMR.
\end{proof}

\subsection{Equal Action Word Syntax Embedding}
\label{sec:equal_action_word_syntax_embedding}
We now prove Lemma~\ref{lemma:scan_task_algorithm_property}.

\subsubsection{Minimization}
\label{section:minimization}
We analyze the minimum entropy for syntax and semantic representations, achievable simultaneously. We show that in such cases, action words have equal syntax embeddings. We use the following proof technique.

\begin{lemma}[Minimum Point Membership]
\label{lemma:minimum_point_membership}
Given: 
\begin{itemize}
    \item a set: $X$,
    \item a function $f: X \to \mathbb{R}$,
    \item an indicator function $g: X \to \{0, 1\}$,
\end{itemize}
and defining:
\begin{itemize}
    \item $X^* = \{x^* : f(x^*) = \min_{x \in X} f(x)\}$,
    \item $X_i = \{x \in X: g(x) = i\}, \forall i \in \{0, 1\}$,
\end{itemize}
we have:
\begin{align*}
    \text{if} \quad & \exists x_1 \in X_1, \forall x_0 \in X_0: f(x_1) < f(x_0), \\
    \text{then} \quad & \forall x^* \in X^*: g(x^*) = 1.
\end{align*}
\end{lemma}
\begin{proof}
$\forall x^* \in X^*:$
\begin{align*}
    \because \; & \forall x_0 \in X_0: f(x^*) \leq f(x_1) < f(x_0) \\
    \therefore \; & x^* \not\in X_0
    \implies x^* \in X_1
    \implies g(x^*) = 1
\end{align*}
\end{proof}

\begin{replemma}{lemma:scan_task_algorithm_property}[SCAN Task Algorithm Property]
With the algorithm in Section~\ref{section:scan_algorithm}, when correct training predictions hold and the entropies of syntax and semantic representations are minimized, all action words have equal hypothesis syntax embeddings.
\end{replemma}
\begin{proof}
We first construct a target syntax embedding that has equal action word embeddings and enables correct training predictions (Lemma~\ref{lemma:target_syntax_embedding_property}). We then verify that there is no alternative syntax embedding with:
\begin{itemize}
    \item non-equal action word embeddings,
    \item correct training predictions, and
    \item entropy no larger than that of the target syntax embedding (Lemma~\ref{lemma:alternative_syntax_embedding}).
\end{itemize}
We apply Lemma~\ref{lemma:minimum_point_membership} with:
\begin{itemize}
    \item $X$ = set of embeddings that yield correct training predictions;
    \item $f$ = entropy; and
    \item $g$ = indicator function, where $g=1$ if all action words have equal syntax representations, and $g=0$ otherwise.
\end{itemize}
The target syntax embedding satisfies the condition; hence, any entropy‑minimizing embedding has $g=1$, i.e., all action words have equal syntax representations.
\end{proof}

\subsubsection{Minimum Semantic Embedding}
We investigate a semantic embedding with minimum entropy.
\begin{definition}[Minimum Semantic Embedding]
\label{definition:minimum_semantic_embedding}
Action words, direction words, and \texttt{<eos>} have distinct embeddings, and all other words share the same embedding as the most frequent action word, direction word, or \texttt{<eos>}.
\end{definition}

\begin{lemma}[Minimum Semantic Embedding and Minimum Entropy]
\label{lemma:minimum_semantic_embedding_and_minimum_entropy}
Under the condition of correct training predictions, the minimum semantic embedding (Definition~\ref{definition:minimum_semantic_embedding}) achieves the minimum entropy.
\end{lemma}
\begin{proof}
Action words and \texttt{<eos>} have distinct semantic embeddings due to single-word samples such as \texttt{run} vs. \texttt{walk}, because action outputs can be generated only from action words, and the output token \texttt{EOS} only from the input token \texttt{<eos>}.

Direction words also have distinct semantic embeddings due to action–direction samples such as \texttt{run left} vs. \texttt{run right}. The attention for the direction output cannot be on action words or \texttt{<eos>} because they do not have semantic embeddings of direction words; therefore, the attentions are on direction words. Hence, the semantic embeddings for direction words differ from those for action words or \texttt{<eos>}, and are distinct among themselves.

Thus, all action words, direction words, and \texttt{<eos>} must have distinct semantic embeddings, and they cover all output primitives. By the definition of entropy, the minimum is attained when the most frequent embedding is assigned to the remaining words (\texttt{and, after, twice, opposite, around, thrice, turn}). Therefore, the minimum semantic embedding (Definition~\ref{definition:minimum_semantic_embedding}) achieves the minimum entropy.
\end{proof}
Note that the conclusion of this lemma does not depend on the syntax word embeddings.

\subsubsection{Target Syntax Embedding}
\label{sec:target_syntax_embedding}
We define standard and target syntax embeddings and analyze their properties. Since \texttt{<eos>} always appears at the end of the input and only there, it does not affect the syntactic structure, so we exclude it from the subsequent discussion.

\paragraph{Standard Syntax Embedding}
\begin{definition}[Standard Syntax Embedding]
\label{definition:standard_syntax_embedding}
A standard syntax embedding satisfies:
\begin{itemize}
    \item all action words share the same embedding;
    \item all direction words share the same embedding; and
    \item any two words not covered by the above two rules have distinct embeddings.
\end{itemize}
\end{definition}
For example, we can produce a standard syntax embedding by replacing action words with 0 and direction words with 1:
\begin{center}
    \begin{tabular}{lccccc}
         Input: & \texttt{run} & \texttt{twice} & \texttt{and} & \texttt{walk} & \texttt{left} \\
         Standard: & 0 & \texttt{twice} & \texttt{and} & 0 & 1
    \end{tabular}
\end{center}

\begin{lemma}[Standard Syntax Embedding Property]
\label{lemma:standard_syntax_embedding_property}
A standard syntax embedding (Definition~\ref{definition:standard_syntax_embedding}) enables correct training predictions when combined with the minimum semantic embedding (Definition~\ref{definition:minimum_semantic_embedding}).
\end{lemma}
\begin{proof}
By construction, each input can have only one template, and it does not depend on the actual action or direction words in the input. Hence, the template can be identified without knowing concrete action or direction words. Also, by Definition~\ref{definition:minimum_semantic_embedding}, the semantic embeddings are distinct for the words corresponding to the outputs. Therefore, the standard syntax embedding allows correct training predictions.
\end{proof}

\begin{lemma}[Preservation under Bijection]
\label{lemma:preservation_under_bijection}
If a syntax embedding has sentence syntax embeddings bijective with those of the standard syntax embedding (Definition~\ref{definition:standard_syntax_embedding}), it enables correct training predictions when combined with the minimum semantic embedding (Definition~\ref{definition:minimum_semantic_embedding}).
\end{lemma}
\begin{proof}
The standard syntax embedding enables correct training predictions (Lemma~\ref{lemma:standard_syntax_embedding_property}). A bijection allows a network to learn the corresponding mapping, preserving correct training predictions.
\end{proof}

\paragraph{Target Syntax Embedding}
\begin{definition}[Target Syntax Embedding]
\label{definition:target_syntax_embedding}
The target syntax embedding assigns equal embeddings to all words within the same row and distinct embeddings to words from different rows of the following table:
\begin{center}
\begin{tabular}{l}
    \texttt{jump}, \texttt{run}, \texttt{walk}, \texttt{look}, \texttt{left}, \texttt{right}, \texttt{around} \\
    \texttt{turn}, \texttt{twice}, \texttt{opposite} \\
    \texttt{and}, \texttt{thrice} \\
    \texttt{after}
\end{tabular}
\end{center}
\end{definition}

\begin{lemma}[Target Syntax Embedding Property]
\label{lemma:target_syntax_embedding_property}
The target syntax embedding (Definition~\ref{definition:target_syntax_embedding}) has equal action syntax embeddings and, when combined with the minimum semantic embedding (Definition~\ref{definition:minimum_semantic_embedding}), enables correct training predictions.
\end{lemma}
\begin{proof}
By Definition~\ref{definition:target_syntax_embedding}, all action words share the same syntax embedding. We verified by writing a computer program that there exists a bijection between the sentence syntax embeddings of the target syntax embedding and those of the standard syntax embedding (Definition~\ref{definition:standard_syntax_embedding}). The code is in the repository. Lemma~\ref{lemma:preservation_under_bijection} then guarantees correct training predictions.
\end{proof}

\subsubsection{Alternative Syntax Embedding}
We use the minimum semantic embedding (Definition~\ref{definition:minimum_semantic_embedding}) and the target syntax embedding (Definition~\ref{definition:target_syntax_embedding}). We now show that no alternative syntax embedding with correct training predictions, lower or equal entropy, and non‑equal action embeddings exists.
Since the semantic embedding has the minimum entropy for the minimum semantic embedding (Lemma~\ref{lemma:minimum_semantic_embedding_and_minimum_entropy}), we only focus on the entropy of the syntax embedding.

\begin{lemma}[Alternative Syntax Embedding]
\label{lemma:alternative_syntax_embedding}
There is no syntax embedding that simultaneously satisfies:
\begin{itemize}
    \item action words do not all have equal syntax embeddings;
    \item the embedding enables correct training predictions; and
    \item its entropy is not larger than that of the target syntax embedding.
\end{itemize}
\end{lemma}
\begin{proof}
We wrote a computer program to enumerate all possible embeddings and verified that no such alternative exists.
The code is in the repository.
The enumeration uses the following techniques to rule out potential counterexamples:

\paragraph{Semantic conflict} 
Some syntax embeddings must be distinguished to avoid training prediction errors.
For example, \texttt{and} = \texttt{after} and \texttt{run} = \texttt{walk} should not occur simultaneously.
Consider the sentences:
\begin{center}
\begin{tabular}{ll}
     \texttt{run and walk}, & \texttt{run after walk}.
\end{tabular}
\end{center}
If \texttt{and} = \texttt{after} and \texttt{run} = \texttt{walk} both hold for syntax embedding, the two sentences produce the same attention sequence, but their correct outputs differ (\texttt{RUN WALK} vs. \texttt{WALK RUN}). Since only the second word is a non‑action word, at least one attention must be on an action word, causing a wrong prediction in at least one of the samples.
For example, suppose the first attention is on \texttt{run}; then the second sample has \texttt{RUN} as the first output, so the prediction is wrong.

\paragraph{Output length conflict}
Identical sentence syntax embeddings produce the same output length for training samples.
Table~\ref{table:output_length_conflict} lists rules where equal syntax embeddings lead to different output lengths: conflicts between training samples.
\begin{table}[!ht]
    \centering
    \caption{
        Output length conflict. If two words (first two columns) have equal syntax embeddings, then the sentence syntax embeddings become equal (third column), but the output lengths differ (last two columns).
    }
    \begin{tabular}{lllcc}
        \multicolumn{2}{c}{Word} & Input & \multicolumn{2}{c}{Length} \\
        \hline
        \texttt{run} & \texttt{turn} & \texttt{[]} \texttt{left} & 2 & 1 \\
        \texttt{twice} & \texttt{thrice} & \texttt{run} \texttt{[]} & 2 & 3 \\
        \texttt{opposite} & \texttt{around} & \texttt{turn} \texttt{[]} \texttt{left} & 2 & 4 \\
    \end{tabular}
    \label{table:output_length_conflict}
\end{table}

\paragraph{Symmetry}
Permuting non-\texttt{jump} action words (\texttt{run, walk, look}) together with their corresponding outputs does not change the training data. Hence, embeddings under such permutations are equivalent, and their enumerations can be merged. The same holds for direction words (\texttt{left, right}).
\end{proof}

\section{COGS Task}
\label{section:cogs_task}
We discuss the COGS dataset~\cite{kim-linzen-2020-cogs} in the context of the neuro-symbolic architecture Compositional Program Generator (CPG), which achieves correct generalization on a subset of the problem~\cite{klinger2023compositional}. We discuss how CPG's design aligns with AS-UMR (Definition~\ref{definition:aligned_structure-unambiguous_minimized_representation}) in terms of its three conditions: structural alignment, unambiguous representation, and minimized representation.

\subsection*{Structural Alignment}
CPG explicitly uses a context-free grammar of the input language and a parser (Lark) to produce a parse tree for each input sentence. The model then recursively applies type-specific semantic functions, which correspond to the components defined in this paper, for each grammar rule. This is precisely a computational graph whose structure mirrors the hierarchical composition defined by the grammar. The output is generated by composing primitive meanings (dictionary) bottom-up via rules.

\subsection*{Unambiguous and Minimized Representations}
Unambiguous and minimized representations capture the notion of encoding adequate but not excessive information. While an unambiguous representation is data-dependent, CPG enforces a minimized representation by limiting programs to two types (copy and substitution), and it is further strengthened by discrete symbolic output (via sampling).

\end{document}